\documentclass[twoside,11pt]{article} 
\usepackage{jmlr_arxiv}

\usepackage{amsmath,amsfonts,bm}

\def\Figref#1{Figure~\ref{#1}}

\def\Secref#1{Section~\ref{#1}}
\def\Appref#1{Appendix~\ref{#1}}

\def\eqref#1{equation~\ref{#1}}

\def\1{\bm{1}}

\def\ra{{\textnormal{a}}}
\def\rb{{\textnormal{b}}}

\def\rd{{\textnormal{d}}}

\def\rf{{\textnormal{f}}}

\def\ro{{\textnormal{o}}}
\def\rp{{\textnormal{p}}}

\def\ry{{\textnormal{y}}}
\def\rz{{\textnormal{z}}}

\def\rvm{{\mathbf{m}}}

\def\rvo{{\mathbf{o}}}

\def\rvv{{\mathbf{v}}}
\def\rvw{{\mathbf{w}}}
\def\rvx{{\mathbf{x}}}

\def\rvz{{\mathbf{z}}}

\def\rmR{{\mathbf{R}}}

\def\rmT{{\mathbf{T}}}

\DeclareMathAlphabet{\mathsfit}{\encodingdefault}{\sfdefault}{m}{sl}
\SetMathAlphabet{\mathsfit}{bold}{\encodingdefault}{\sfdefault}{bx}{n}

\def\gC{{\mathcal{C}}}
\def\gD{{\mathcal{D}}}

\def\gQ{{\mathcal{Q}}}

\def\gS{{\mathcal{S}}}

\newcommand{\E}{\mathbb{E}}
\newcommand{\Ls}{\mathcal{L}}
\newcommand{\R}{\mathbb{R}}

\newcommand{\KL}{D_{\mathrm{KL}}}

\usepackage{amsmath,amsfonts,bm}

\newcommand{\paren}[1]{{( #1 )}}
\newcommand{\bigparen}[1]{{\left( #1 \right)}}

\newcommand{\AVP}{{\text{APP}}}
\newcommand{\GQN}{{\text{GQN}}}
\newcommand{\MLP}{{\text{MLP}}}

\newcommand{\ST}[1]{{\mathcal{ST}\paren{#1}}}
\newcommand{\IST}[1]{{\mathcal{ST}^{-1}\paren{#1}}}

\newcommand{\N}[1]{{\mathcal{N}\paren{#1}}}

\newcommand{\bigN}[1]{{\mathcal{N}\bigparen{#1}}}

\newcommand{\enc}{{\mathrm{enc}}}
\newcommand{\dec}{{\mathrm{dec}}}

\newcommand{\bg}{{\mathrm{bg}}}

\newcommand{\pres}{{\mathrm{pres}}}
\newcommand{\where}{{\mathrm{where}}}
\newcommand{\what}{{\mathrm{what}}}

\newcommand{\ctr}{{\mathrm{center}}}
\newcommand{\scale}{{\mathrm{scale}}}
\newcommand{\depth}{{\mathrm{depth}}}
\newcommand{\pos}{{\mathrm{pos}}}
\newcommand{\att}{{\mathrm{att}}}
\newcommand{\nbr}{{\mathrm{neighbor}}}

\newcommand{\bpsi}{\boldsymbol{\psi}}
\newcommand{\upd}{\textup{d}}

\usepackage[utf8]{inputenc} 

\usepackage{booktabs}       
\usepackage{nicefrac}       
\usepackage{microtype}      
\usepackage{microtype}
\usepackage{graphicx}
\usepackage{algorithm}
\usepackage{algorithmic}
\usepackage[inline]{enumitem}
\usepackage{enumitem}
\usepackage{wrapfig}
\usepackage{lipsum}
\usepackage{accents}
\usepackage{array}
\usepackage{mathtools}
\usepackage{mathrsfs}
\usepackage{placeins}
\usepackage{multirow}
\usepackage{tikz}
\usepackage{pgfplots}
\pgfplotsset{compat=1.16}
\usepackage{pgfplotstable}
\usepgfplotslibrary{groupplots}
\usepackage{adjustbox}
\usepackage{url}

\firstpageno{1}

\begin{document}

\title{ROOTS: Object-Centric Representation and Rendering of 3D Scenes}

\author{\name Chang Chen\thanks{Both authors contributed equally.} \email chang.chen@rutgers.edu \\
        \addr Department of Computer Science\\
        Rutgers University\\
        Piscataway, NJ 08854, USA
        \AND
        \name Fei Deng$^*$ \email fei.deng@rutgers.edu \\
        \addr Department of Computer Science\\
        Rutgers University\\
        Piscataway, NJ 08854, USA
        \AND
        \name Sungjin Ahn \email sungjin.ahn@rutgers.edu \\
        \addr Department of Computer Science and Center for Cognitive Science\\
        Rutgers University\\
        Piscataway, NJ 08854, USA}

\editor{}

\maketitle

\begin{abstract}
A crucial ability of human intelligence is to build up models of individual 3D objects from partial scene observations. Recent works achieve object-centric generation but without the ability to infer the representation, or achieve 3D scene representation learning but without object-centric compositionality. Therefore, learning to represent and render 3D scenes with object-centric compositionality remains elusive. In this paper, we propose a probabilistic generative model for learning to build modular and compositional 3D object models from partial observations of a multi-object scene. The proposed model can (i) infer the 3D object representations by learning to search and group object areas and also (ii) render from an arbitrary viewpoint not only individual objects but also the full scene by compositing the objects. The entire learning process is unsupervised and end-to-end. In experiments, in addition to generation quality, we also demonstrate that the learned representation permits object-wise manipulation and novel scene generation, and generalizes to various settings. Results can be found on our project website: \url{https://sites.google.com/view/roots3d}
\end{abstract}


\section{Introduction}
At the core of human learning is the ability to build up mental models of the world along with the growing experience of our life. In building such models, a particularly important aspect is to factorize underlying structures of the world such as objects and their relationships. This ability is believed to be crucial in enabling various advanced cognitive functions in human-like AI systems~\citep{lake2017building} such as systematic generalization~\citep{bahdanau2018systematic,van2019perspective}, reasoning~\citep{bottou2014machine}, and causal learning~\citep{schlkopf2019causality,peters2017elements}. While humans seem to learn such object-centric  representations~\citep{objectfiles,rolls2005object,hood2009origins,von1985object,martin2007representation,hoydal2019object} in a 3D-aware fashion through partial observations of scenes without supervision, in machine learning this problem has only been tackled for simple 2D fully-observable images~\citep{air,space,spair,nem,iodine,slotattention,monet,genesis}. Therefore, the more challenging yet realistic setting of learning 3D-aware object-centric representation of 3D space from partial observations has remained elusive.

Regarding this, there have been a number of recent approaches that can (only) \textit{generate} 3D scene images via object-centric compositional rendering~\citep{blockgan,van2020investigating,ehrhardt2020relate}. However, none of the existing models provide the crucial ability of the reverse that we seek in this paper: the object-centric inverse graphics, i.e., learning object-centric 3D representations from partial observations. In learning representations and rendering of 3D scenes, GQN~\citep{gqn} and its variants~\citep{cgqn,tobin2019geometry,snp,asnp} are the most close to our work. However, the 3D representations inferred by these models provide only scene-level representation without explicit object-centric decomposition.

In this paper, we tackle the problem of learning to build modular and compositional 3D object models from partial scene images. Our proposed model, ROOTS ($\underaccent{\bar}{\text{R}}$epresentation and Rendering of $\underaccent{\bar}{\text{O}}$bject-$\underaccent{\bar}{\text{O}}$riented $\underaccent{\bar}{\text{T}}$hree-D $\underaccent{\bar}{\text{S}}$cenes), is able to decompose partial observations into objects, group them object-wise, and build a modular compositional 3D representation at the level of individual objects. Such representation also enables compositional rendering. As our object model provides object-wise 3D rendering from arbitrary viewpoint, we can also render the entire scene from arbitrary viewpoints by first rendering individual objects and then compositing them according to the scene layout. In particular, this enables a novel \textit{nested autoencoder architecture} in which we can reuse the GQN model as a internal autoencoder module for object modeling, making the model simpler. The entire process is unsupervised and end-to-end trainable. We demonstrate the above capabilities of our model on simulated 3D scenes with multiple objects. We evaluate our model in terms of generation quality, structure accuracy, generalization ability, and downstream task performance. We also showcase that by manipulating the scene layout, we can generate scenes with many more objects than typical of the training regime.




\section{Preliminary: Generative Query Networks}
\label{sec:gqn}

The Generative Query Network (GQN) is a latent variable model for learning to represent and render 3D scenes. Given a set of context images and viewpoints, it learns a \textit{3D-viewpoint-steerable representation} (in short, \textit{3D representation} throughout this paper) in the sense that any target image viewed from an arbitrary viewpoint can be generated from the representation. We note that such 3D representations are different from and more challenging to learn than \textit{2D representations} that only model the scene from a single viewpoint. Recent advances in unsupervised object-centric representation learning \citep{air,space,spair,nem,iodine,slotattention,monet,genesis} mostly require the representation to model only a single 2D image. Therefore, these methods can only learn 2D representations, even if the 2D image is a view of an underlying 3D scene.  




More formally, consider an agent navigating a 3D environment (called a scene) and collecting $K$ pairs of image $\rvx_c$ and the corresponding viewpoint $\rvv_c$ for $c=1, 2, \dots, K$. This collection is called \textit{context} $\gC = \{(\rvx_c,\rvv_c)\}_{c=1}^K$. GQN learns a scene-level 3D representation $\rvz$ by encoding $\gC$, such that the target image $\hat{\rvx}_q$ from an arbitrary query viewpoint $\rvv_q$ can be generated by the decoder $\hat{\rvx}_q = \GQN_\dec(\rvz,\rvv_q)$. The generative process can be written as:
\begin{align*}
    p(\rvx_q \mid \rvv_q, \gC) = \int p(\rvx_q \mid \rvz, \rvv_q) \, p(\rvz \mid \gC) \, \upd \rvz \ .
\end{align*}
The prior encoder $p(\rvz \!\mid \! \gC) = \GQN_\enc(\gC)$ first obtains an order-invariant encoding (e.g., a sum encoding) $\bm{r}_{\gC}$ of context $\gC$, and then uses ConvDRAW~\citep{gregor2016towards} to autoregressively sample $\rvz$ from $\bm{r}_{\gC}$. The decoder $\GQN_\dec(\rvz,\rvv_q)$ uses a deterministic version of ConvDRAW to render the target image $\hat{\rvx}_q$ from $\rvz$, and $p(\rvx_q \mid \rvz, \rvv_q)$ is often modeled as a Gaussian distribution $\N{\hat{\rvx}_q, \sigma^2\1}$ with $\sigma$ being a hyperparameter. Since computing the posterior distribution $p(\rvz \!\mid \! \rvx_q,\rvv_q,\gC)$ is intractable, GQN uses variational inference for posterior approximation and is trained by maximizing its evidence lower bound. 
Backpropagation through random variables is done by the reparameterization trick~\citep{vae,vae_rezende}. 

Note that the model described above is actually a more consistent version of the GQN named CGQN~\citep{cgqn}. In the original GQN~\citep{gqn}, the latent $\rvz$ is also conditioned on $\rvv_q$, i.e., $p(\rvz \!\mid \! \rvv_q, \gC)$, and rendering is query-agnostic, i.e., $p(\rvx_q \!\mid \! \rvz)$, leading to potential inconsistency across multiple query viewpoints. Throughout the paper, we use the abbreviation GQN to refer to the general GQN framework embracing both GQN and CGQN.

\section{ROOTS}

GQN represents a multi-object 3D scene as a single vector without learning explicit object-wise decomposition. Hence, it cannot entertain the potential and various advantages of object-centric representations. To resolve this limitation, we propose ROOTS, a probabilistic generative model that learns to represent and render 3D scenes via composition of object-centric 3D representations in a fully unsupervised and end-to-end trainable way. This problem has never been tackled, and it is highly challenging because not only can an object be unobservable from certain viewpoints, but also the appearance, position, pose, size, and occlusion of an object can vary significantly across the context images. The premise of our approach to tackling this challenge is that: if we can collect the local regions corresponding to a specific object across the context images, 
then we can reuse GQN on those filtered local observations to learn the 3D representation for that object.

\begin{figure}[t]
\centering
\includegraphics[width=\textwidth]{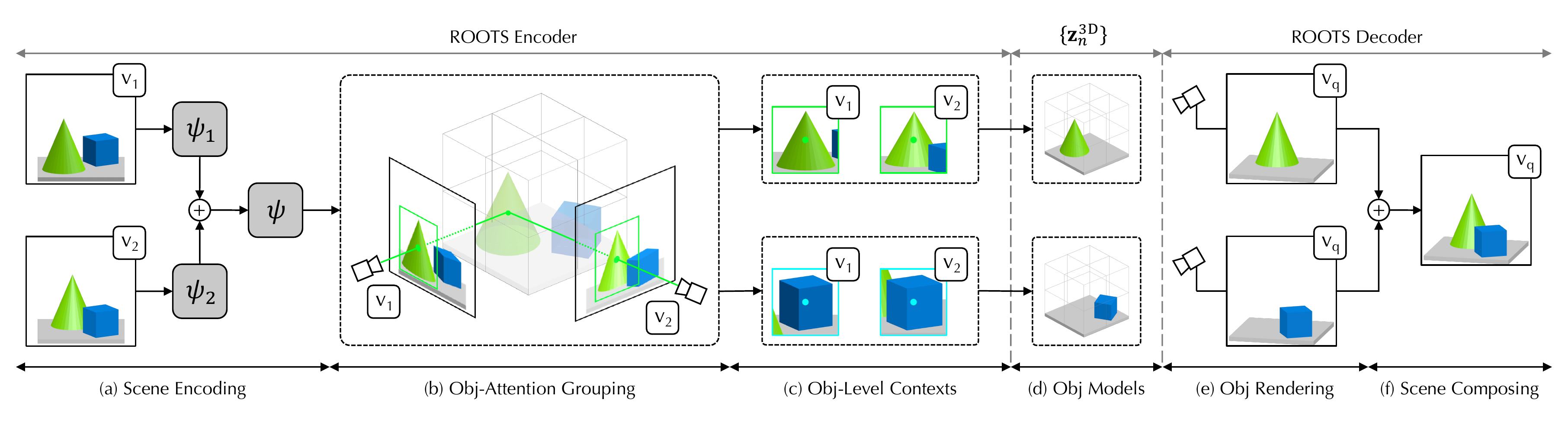}
\caption{Overview of ROOTS pipeline.
{\em{ROOTS encoder (a - c)}}: (a) Context observations are encoded and aggregated into a scene-level representation $\bpsi$. (b) $\bpsi$ is reorganized into a feature map of the 3D space, from which 3D center positions are inferred for each object. By applying perspective projection to the inferred 3D center positions, we identify image regions for each object across viewpoints. (c) Object regions are cropped and grouped into object-level contexts. {\em{Object models (d)}}: The object-level contexts allow us to obtain the 3D appearance representation of each object through an object-level GQN. {\em{ROOTS decoder (e - f)}}: To render the full scene for a given query viewpoint, we composite the rendering results of individual objects.}
\label{fig:overview}
\end{figure}

To this end, we propose the following approaches. First, ROOTS has a \textit{nested autoencoder architecture}, one autoencoder at scene-level and the other at object-level. Further, the scene-level autoencoder is constructed by the composition of the object-level autoencoders. For the scene-level encoding, the model encodes the context set to a 3D spatial structure of the scene and infers the 3D position of each object existing in the 3D space. Given the inferred 3D position of objects, we then propose a method, called \textit{Attention-by-Perspective-Projection} to efficiently find and attend the local regions, corresponding to a specific object, across all the context images. This grouping allows us to construct a new object-level context set containing only a specific object and thus to reuse the standard GQN encoder as an in-network module for object-level 3D-aware encoding of the object appearance. Scene-level decoding is also composed by object-level decoding and background-decoding. We decode the appearance representation of each object using the object-level GQN decoder, and place the decoded images in the target image by mapping the 3D positions to the 2D positions in the target image. Together with background rendering, we can complete the rendering of a scene image. See \Figref{fig:overview} for an overview of ROOTS pipeline.

\subsection{ROOTS Encoder} \label{sec:encoder}

The goal of ROOTS encoder is to infer the \textit{3D object models} from scene-level context observations $\gC = \{(\rvx_c, \rvv_c)\}_{c=1}^{K}$ but without any object-level supervision. Each object model consists of the 3D representation of an object, fully disentangled into its 3D position and 3D appearance. The modularity and compositionality of these object models allow them to be collected from multiple scenes, and then reconfigured to generate novel scenes that are out of the training distribution.

To infer the 3D object models, it is imperative that the encoder should be properly structured. In particular, we find in our experiments that directly inferring object models from an order-invariant encoding of $\gC$ would fail, potentially because the lack of proper structure prohibits learning and optimization. To solve this problem, we extract object regions from each scene image $\rvx_c$ and group them into object-level contexts, which provide more relevant information for inferring the 3D appearance of each object. We call this grouping process object-attention grouping.

For 2D fully observable images, extracting object regions can be solved by recent 2D scene decomposition methods \citep{air,spair,space}. However, in our 3D and partially observed setting, it remains a challenge to efficiently group together the regions that correspond to the same object across viewpoints. One naive approach is to find the best among all possible groupings, but its time complexity is exponential in the number of viewpoints. Another possible way is to treat the extracted object regions in one viewpoint as anchors, and match regions from other viewpoints to one of these anchors, by computing pairwise matching scores. The time complexity is quadratic in the number of objects. By contrast, our proposed object-attention grouping scales linearly in both the number of viewpoints and the number of objects. The key idea is to first infer the center position of each object in 3D coordinates. This allows us to use perspective projection \citep{hartley2003multiple} from 3D to 2D to efficiently locate the same object across different context images.

We develop a scene encoder (\Secref{sec:GVFM}) to infer the object positions, describe in more detail the object-attention grouping in \Secref{sec:AVP}, and use an object-level GQN encoder (\Secref{sec:OGQN}) to infer the object appearance.

\subsubsection{Scene Encoder} \label{sec:GVFM}

The goal of the scene encoder is to infer the 3D center position of each object in world coordinates---the same coordinate system where the camera viewpoints $\rvv_c$ are measured. We assume that the objects resides in a bounded 3D space.
The scene encoder partitions this bounded 3D space into a rectangular cuboid of ${N_{\mathrm{max}} = N_x \times N_y \times N_z}$ cells. For each cell $(i,j,k)$, we infer a Bernoulli variable ${\rz_{ijk}^\pres \in \{0,1\}}$ that is $1$ if and only if the cell contains the center of an object (note that the full appearance volume of an object need not be contained within the cell). We also infer a continuous variable ${\rvz_{ijk}^\where \in \R^3}$ that, when $\rz_{ijk}^\pres = 1$, specifies the center position of the object in the 3D world coordinates. Here ${\rvz_{ijk}^\where}$ is constrained to be within the boundary of cell $(i,j,k)$. This prior on cell-wise object preference helps efficient training and obviates the need for expensive autoregressive processing~\citep{space}.


In the above scene encoding, each cell handles  one or no object. In actual implementation, however, the partition is soft, meaning neighboring cells can have some overlap. Hence, when a cell does contain more than one object, the scene encoder can learn to distribute them to adjacent cells. We can also simply increase the resolution of the 3D partitioning. A similar 2D version of this approach has been used in SPAIR \citep{spair} and SPACE \citep{space}, showing impressive decomposition of 2D scenes into dozens of objects. 

Specifically, to infer ${\{(\rz_{ijk}^\pres, \rvz_{ijk}^\where)\}}$ from the context observations $\gC$, we encode $\gC$ into a Geometric Volume Feature Map (GVFM) $\bm{r} \in \R^{N_x \times N_y \times N_z \times d}$, yielding a $d$-dimensional feature vector $\bm{r}_{ijk}$ for each cell. Then, $\{(\rz_{ijk}^\pres, \rvz_{ijk}^\where)\}$ can be computed in parallel for all cells by a neural network $f_{\pres, \where}$:
\begin{equation*} \label{eqn:z_where}
    p(\rz_{ijk}^\pres, \rvz_{ijk}^\where \mid \gC) = f_{\pres, \where}(\bm{r}_{ijk}^\nbr)\ ,
\end{equation*}
where ${\bm{r}_{ijk}^\nbr}$ includes the feature vectors of cell $(i,j,k)$ and its neighboring cells, allowing inter-object relations to be taken into consideration.

While GVFM may seem similar to the grid cells used in SPAIR and SPACE, there are fundamental differences. As a feature map of the 3D space, GVFM must \emph{aggregate} information from multiple partial 2D observations and \emph{reorganize} it in an object-wise fashion. This is in contrast to the 2D feature map learned by grid cells which have a natural alignment with the single fully-observed 2D image. Therefore, we obtain GVFM in two steps. First, we compute an order-invariant summary $\bpsi$ of $\gC$ as the summation over encodings of individual context observations:
\begin{equation*}
    \bpsi = {\textstyle\sum}_{c=1}^{K}\,\bpsi_c = {\textstyle\sum}_{c=1}^{K}\,f_{\bpsi}(\rvx_c, \rvv_c)\ ,
\end{equation*}
where $f_{\bpsi}$ is a learned encoding network. Second, we apply a 3D transposed convolution over $\bpsi$ to turn the sum of 2D image representations $\bpsi$ into 3D spatial representation $\bm{r}$ where individual $\bm{r}_{ijk}$ slots contains object-specific information:
\begin{equation*}
    \bm{r} = \mathrm{ConvTranspose3D}(\bpsi)\ .
\end{equation*}

\subsubsection{Object-Attention Grouping} \label{sec:AVP}

Object-attention grouping aims to identify image regions that correspond to the same object across different observation images. This is crucial in obtaining object-wise 3D appearance representations. More precisely, for each object $n$ present in the scene and each context image $\rvx_c$, we seek a 2D bounding box capturing object $n$ in $\rvx_c$. The bounding box is parameterized by its center position and scale (width and height), denoted $(\rvo_{n,c}^\ctr, \rvo_{n,c}^\scale)$. Notice that here each object index $n$ corresponds to a distinct cell index $(i,j,k)$ with $\rz_{ijk}^\pres = 1$.


Our key observation is that inferring the 3D object center positions in the first step allows us to solve object-attention grouping by using perspective projection. We call this Attention-by-Perspective-Projection (APP). Assuming that the projection operation takes constant time, the time complexity of APP is linear in both the number of objects and the number of viewpoints.


\textbf{Attention-by-Perspective-Projection (APP).}
Let us focus on object $n$ and find its 2D bounding box in $\rvx_c$. We first analytically compute its 2D center position ${\rvo_{n,c}^\ctr} \in \R^2$ in $\rvx_c$ and its distance from the camera, denoted ${\ro_{n,c}^\depth} \in \R$, by applying perspective projection to its 3D center position ${\rvz_{n}^\where}$:
\begin{equation*} \label{eqn:s_where}
    [\rvo_{n,c}^\ctr, \ro_{n,c}^\depth]^\top = \AVP_\pos(\rvz_{n}^\where, \rvv_c)  = \mathrm{normalize} (\rmT_{\mathrm{World}\rightarrow\mathrm{Camera}}(\rvv_c) [\rvz_n^\where, 1]^\top)  \ .
\end{equation*}
Here, $\rvz_n^\where$ is first converted to camera coordinates by the viewpoint-dependent transformation matrix $\rmT_{\mathrm{World}\rightarrow\mathrm{Camera}}(\rvv_c) \in \R^{3 \times 4}$, and then normalized. See \Appref{app:proj} for more details.

To compute the 2D bounding box scale $\rvo_{n,c}^\scale \in \R^2$, one option is to learn a 3D bounding box for object $n$, project its eight vertices onto the image plane, and find the smallest rectangle that covers all eight vertices. Unfortunately, the resulting 2D bounding box will only be tight under specific viewpoints, and we will likely encounter optimization difficulties. Hence, to allow better gradient flow and provide the model with the opportunity to predict tighter 2D bounding boxes, we design $\AVP_\scale$ that implicitly learns the projection:
\begin{equation*} \label{eqn:s_scale}
    p(\rvo_{n,c}^\scale \!\mid\! \rvz_n^\where, \gC) = \AVP_\scale(\rvo_{n,c}^\ctr, \ro_{n,c}^\depth, \bm{r}_n, \rvv_c) = \MLP(\texttt{concat}[\rvo_{n,c}^\ctr, \ro_{n,c}^\depth, \bm{r}_n, \rvv_c]) \ .
\end{equation*}
To work properly, $\AVP_\scale$ should learn to perform the following operation implicitly: to extract 3D scale information from $\bm{r}_n$, make a projection from viewpoint $\rvv_c$, and refine the projection using $\rvo_{n,c}^\ctr$ and $\ro_{n,c}^\depth$.

\subsubsection{Object Encoder} \label{sec:OGQN}

With object-attention grouping, we can decompose the scene-level context $\gC$ into object-level context ${\gC}_n$ for each object $n = 1, 2, \dots, N$, where
\begin{equation*} \label{eqn:num_obj}
    N = {\textstyle\sum}_{ijk}\,\rz_{ijk}^\pres \leq N_{\mathrm{max}}
\end{equation*}
is the total number of objects present in the scene. Specifically, we first use a spatial transformer $\mathcal{ST}$ \citep{spatial_transformer} to differentiably crop object patch $\rvx_{n,c}^\att$ from scene image $\rvx_c$ using $\rvo_{n,c}^\ctr$ and $\rvo_{n,c}^\scale$:
\begin{equation*}
    \rvx_{n,c}^\att = \ST{\rvx_c, \rvo_{n,c}^\ctr, \rvo_{n,c}^\scale}\ .
\end{equation*}
After collecting these patches from all viewpoints, we group them based on the object index $n$ to obtain object-level context
\begin{equation*}
    {\gC}_n = \{(\rvx_{n,c}^\att, \rvv_c, \rvo_{n,c}^\where)\}_{c=1}^{K}\ ,
\end{equation*}
where we include $\rvo_{n,c}^\where = (\rvo_{n,c}^\ctr, \rvo_{n,c}^\scale, \ro_{n,c}^\depth)$ to provide information complementary to ${\rvx_{n,c}^\att}$. 
The object-level context allows us to use an object-level GQN encoder
\begin{equation*} \label{eqn:z_what}
    p(\rvz_n^\what \mid {\gC}_n) = \GQN_\enc({\gC}_n)
\end{equation*}
to obtain independent and modular object-level 3D appearance $\rvz_n^\what$ for each object $n$.
A summary of ROOTS encoder is provided in \Appref{app:code}.

\subsection{ROOTS Decoder} \label{sec:decoder}

Given partial observations of a multi-object 3D scene, ROOTS not only learns to infer the 3D object models, but also learns to render them independently and individually from arbitrary viewpoints. The full scene is also rendered from arbitrary query viewpoints by compositing object rendering results. By collecting and re-configuring the inferred object models, ROOTS can easily generate novel scenes that are out of the training distribution.


\textbf{Object Renderer.}
For each object $n$, given its 3D appearance representation $\rvz_n^\what$ and a query viewpoint $\rvv_q$, ROOTS is able to generate a 4-channel (RGB+Mask) image ${\rvo_{n,q}^\what}$ depicting the object's 2D appearance when viewed from $\rvv_q$. This is achieved by an object-level GQN decoder:
\begin{equation*} \label{eqn:ogqn_dec}
    \rvo_{n,q}^\what = \GQN_\dec(\texttt{concat}[\rvz_n^\what, \bm{r}_n^\att], \rvv_q)\ .
\end{equation*}
Here, $\bm{r}_n^\att$ is an order-invariant summary of object-level context ${\gC}_n$:
\begin{equation*}
    \bm{r}_n^\att = {\textstyle\sum}_{c=1}^{K}\,f_{\att}(\rvx_{n,c}^\att, \rvv_c, \rvo_{n,c}^\where)\ ,
\end{equation*}
where $f_{\att}$ is a learnable encoding network.

\textbf{Scene Composer.}
The final scene image $\hat{\rvx}_q$ corresponding to query $\rvv_q$ is obtained by superimposing layers of object-wise images with proper masking. For this, we first use an inverse spatial transformer $\mathcal{ST}^{-1}$ \citep{spatial_transformer} to differentiably place each object at the right position in the scene canvas with proper scaling:
\begin{equation*}
     [\hat{\rvx}_{n,q}, \hat{\rvm}_{n,q}] = \IST{\rvo_{n,q}^\what, \rvo_{n,q}^\ctr, \rvo_{n,q}^\scale}\ .
\end{equation*}
Here, the 3-channel image $\hat{\rvx}_{n,q}$ can be regarded as an object-specific image layer containing only object $n$, and the single-channel $\hat{\rvm}_{n,q}$ is the mask for object $n$. The position and scaling parameters are computed by simply reusing the APP module for query viewpoint $\rvv_q$:
\begin{align*} \label{eqn:o_where}
    [\rvo_{n,q}^\ctr, \ro_{n,q}^\depth]^\top &= \AVP_\pos(\rvz_{n}^\where, \rvv_q)\ , \notag \\
    p(\rvo_{n,q}^\scale \mid \rvz_{n}^\where, \rvv_q, \gC) &= \AVP_\scale(\rvo_{n,q}^\ctr, \ro_{n,q}^\depth, \bm{r}_n, \rvv_q)\ .
\end{align*}
We then composite these $N$ image layers into a single image, ensuring that occlusion among objects is properly handled. Similar to previous works \citep{van2018case,spair,monet,iodine,genesis,space}, for each layer $n$, we compute a transparency map
\begin{equation*}
    \bm{\alpha}_{n,q} = \bm{w}_{n,q} \odot \hat{\rvm}_{n,q}\ ,
\end{equation*}
where $\odot$ is pixel-wise multiplication. This masks out occluded pixels of object $n$. To obtain the values of $\{\bm{w}_{n,q}\}_{n=1}^{N}$ at each pixel, we first use $\{\hat{\rvm}_{n,q}\}_{n=1}^{N}$ to find the objects that contain the pixel, and then assign the values based on their relative depth $\{\ro_{n,q}^\depth\}_{n=1}^{N}$. See \Appref{app:alpha} for more details. The final rendered scene $\hat{\rvx}_{q}$ is composited as:
\begin{equation*} \label{eqn:x_q_hat}
    \hat{\rvx}_{q} = {\textstyle\sum}_{n=1}^{N}\,\bm{\alpha}_{n,q} \odot \hat{\rvx}_{n,q}\ .
\end{equation*}
A summary of ROOTS decoder is provided in \Appref{app:code}.

\subsection{Probabilistic Model} \label{sec:full_model}

We now piece things together and formulate ROOTS as a conditional generative model. Given a collection of context observations $\gC = \{(\rvx_c, \rvv_c)\}_{c=1}^{K}$ of a multi-object scene, ROOTS learns to infer the number of objects, denoted $N$, the 3D object model $\rvz_n^{\mathrm{3D}} = (\rvz_n^\where, \rvz_n^\what)$ for each object $n = 1, 2, \dots, N$, and the 2D representation $\rvo_{n,\gC}^{\mathrm{2D}} = \{\rvo_{n,c}^\where\}_{c=1}^{K}$ collected for each object $n$ from all context viewpoints. In addition, ROOTS also learns a background representation $\rvz^\bg$ through a scene-level GQN encoder:
\begin{equation*}
    p(\rvz^\bg \mid \gC) = \GQN_\enc^\bg(\gC)\ .
\end{equation*}
Using these representations, ROOTS can then generate the target image $\rvx_q$ from an arbitrary query viewpoint $\rvv_q$ of the same scene. During generation, ROOTS also infers the 2D object representation $\rvo_{n,q}^{\mathrm{2D}} = \rvo_{n,q}^\where$ for the query viewpoint. We do not include $\rvo_{n,q}^\what$ here because it is a deterministic variable.

Let $\gQ = \{(\rvx_q, \rvv_q)\}_{q=1}^{M}$ be the collection of queries for the same scene, $\rvx_\gQ = \{\rvx_q\}_{q=1}^{M}$ and $\rvv_\gQ = \{\rvv_q\}_{q=1}^{M}$ be the target images and query viewpoints respectively, and $\gD = \gC \cup \gQ$ be the union of contexts and queries.
To simplify notations, we collect all viewpoint-independent 3D representations into a single variable $\rvz^{\mathrm{3D}}$, including the number of objects, the 3D object models, and the background representation:
\begin{equation*}
    \rvz^{\mathrm{3D}} = (N, \{\rvz_n^{\mathrm{3D}}\}_{n=1}^{N}, \rvz^\bg)\ .
\end{equation*}
We also collect the viewpoint-dependent 2D representations for all objects into a single variable $\rvo_{\gS}^{\mathrm{2D}}$, where the subscript $\gS$ denotes the set of viewpoints. For example,
\begin{gather*}
    \rvo_{\gC}^{\mathrm{2D}} = \{\rvo_{n,\gC}^{\mathrm{2D}}\}_{n=1}^{N}\ , \quad
    \rvo_{q}^{\mathrm{2D}} = \{\rvo_{n,q}^{\mathrm{2D}}\}_{n=1}^{N}\ .
\end{gather*}
The generative process can then be written as:
\begin{equation*}
    p(\rvx_\gQ \!\mid\! \rvv_\gQ, \gC) =
    \iint
    \underbrace{p(\rvz^{\mathrm{3D}}, \rvo_{\gC}^{\mathrm{2D}} \!\mid\! \gC)}_\text{Encoder}
    \prod_{q=1}^{M}
    \underbrace{p(\rvo_{q}^{\mathrm{2D}} \!\mid\! \rvz^{\mathrm{3D}}, \rvv_q, \gC)}_\text{Object Renderer}
    \underbrace{p(\rvx_q \!\mid\! \rvz^{\mathrm{3D}}, \rvo_{\gC \cup q}^{\mathrm{2D}}, \rvv_q, \gC)}_\text{Scene Composer}
    \upd\rvz^{\mathrm{3D}}\upd\rvo_{\gD}^{\mathrm{2D}}\ .
\end{equation*}
The encoder can be further factorized in an object-wise fashion:
\begin{align*}
    p(\rvz^{\mathrm{3D}}, \rvo_{\gC}^{\mathrm{2D}} \!\mid\! \gC) =
    \underbrace{p(\rvz^\bg \!\mid\! \gC)}_\text{Background}
    \underbrace{p(N \!\mid\! \gC)}_\text{Density}
    \prod_{n=1}^N
    \underbrace{p(\rvz_n^\where \!\mid\! \gC)}_\text{Scene Encoder}
    \underbrace{p(\rvo_{n,\gC}^{\mathrm{2D}} \!\mid\! \rvz_n^\where, \gC)}_\text{APP}
    \underbrace{p(\rvz_n^\what \!\mid\! \gC_n)}_\text{Object Encoder}\ ,
\end{align*}
where the object-level context $\gC_n$ is obtained as a deterministic function of $\rvz_n^\where$, $\rvo_{n,\gC}^{\mathrm{2D}}$, and $\gC$. The object renderer can be factorized similarly:
\begin{align*}
    p(\rvo_{q}^{\mathrm{2D}} \!\mid\! \rvz^{\mathrm{3D}}, \rvv_q, \gC) =
    \prod_{n=1}^N
    \underbrace{p(\rvo_{n,q}^{\mathrm{2D}} \!\mid\! \rvz_n^{\where}, \rvv_q, \gC)}_\text{APP}\ .
\end{align*}
The scene composer obtains the full scene from the foreground image $\hat{\rvx}_{q}$ and the background image $\hat{\rvx}_q^\bg$ through alpha compositing:
\begin{equation*}
    p(\rvx_q \!\mid\! \rvz^{\mathrm{3D}}, \rvo_{\gC \cup q}^{\mathrm{2D}}, \rvv_q, \gC) = \bigN{\hat{\rvx}_{q} + (\bm{1} - {\textstyle\sum}_{n=1}^{N}\,\bm{\alpha}_{n,q}) \odot \hat{\rvx}_q^\bg, \ \  \sigma^2\bm{1}}\ ,
\end{equation*}
where $\hat{\rvx}_q^\bg$ is rendered by a scene-level GQN decoder:
\begin{equation*}
    \hat{\rvx}_q^\bg = \GQN_\dec^\bg(\rvz^\bg, \rvv_q)\ ,
\end{equation*}
and $\sigma^2$ is a hyperparameter called pixel-variance.

\subsection{Inference and Learning} \label{sec:training}

Due to the intractability of the log-likelihood $\log p(\rvx_\gQ \!\mid\! \rvv_\gQ, \gC)$, we train ROOTS using variational inference with the following approximate posterior:
\begin{equation*}
    q(\rvz^{\mathrm{3D}}, \rvo_{\gD}^{\mathrm{2D}} \!\mid\! \gD) = q(\rvz^\bg \!\mid\! \gD) \,
    q(N \!\mid\! \gD) \prod_{n=1}^N q(\rvz_n^\where \!\mid\! \gD) \, q(\rvo_{n,\gD}^{\mathrm{2D}} \!\mid\! \rvz_n^\where, \gD) \, q(\rvz_n^\what \!\mid\! \gD_n)\ ,
\end{equation*}
where $\gD_n$ is the object-level context deterministically obtained from $\gD$ using the inferred $\rvz_n^\where$ and $\rvo_{n,\gD}^{\mathrm{2D}}$. The implementation of the approximate posterior is almost the same as ROOTS encoder described in \Secref{sec:encoder}, except that the summary vector $\bpsi$ should now encode the entire $\gD$ instead of only $\gC$. We treat all continuous variables as Gaussian variables, and use reparameterization trick \citep{vae} to sample from the approximate posterior. For discrete variables, we use Gumbel-Softmax trick \citep{jang2016categorical, maddison2016concrete}. The entire model can be trained end-to-end by maximizing the {Evidence Lower Bound (ELBO):}
\begin{align*}
    \Ls &= \E_{q(\rvz^{\mathrm{3D}}, \rvo_{\gD}^{\mathrm{2D}} \mid \gD)} \left[{\textstyle\sum}_{q=1}^{M}\,\log p(\rvx_q \!\mid\! \rvz^{\mathrm{3D}}, \rvo_{\gC \cup q}^{\mathrm{2D}}, \rvv_q, \gC)\right] - \KL[q(\rvz^\bg \!\mid\! \gD) \!\parallel\! p(\rvz^\bg \!\mid\! \gC)] \notag \\
    &- \KL[q(N \!\mid\! \gD) \!\parallel\! p(N \!\mid\! \gC)] - \E_{q(N \mid \gD)} \left[{\textstyle\sum}_{n=1}^{N}\,\KL[q(\rvz_n^{\mathrm{3D}}, \rvo_{n,\gD}^{\mathrm{2D}} \!\mid\! \gD) \!\parallel\! p(\rvz_n^{\mathrm{3D}}, \rvo_{n,\gD}^{\mathrm{2D}} \!\mid\! \rvv_\gQ, \gC)]\right]\ .
\end{align*}

\textbf{Combining with Unconditioned Prior.}
One difficulty in using the conditional prior is that it may not coincide with our prior knowledge of the latent variables. In our experiments, it turns out that biasing the posterior of some variables toward our prior preference helps stabilize the model. We achieve this by introducing additional KL terms between the posterior and \emph{unconditioned} prior (like in VAEs, \citealt{vae,betavae}) to the ELBO. Specifically, the model is trained by maximizing:
\begin{align*}
    \tilde{\Ls} &= \Ls - \gamma\KL[q(N \!\mid\! \gD) \!\parallel\! {\mathrm{Geom}}(\rho)] - \E_{q(N \mid \gD)} \left[{\textstyle\sum}_{n=1}^{N}\,\KL[q(\rvz_n^\where \!\mid\! \gD) \!\parallel\! \N{\bm{0}, \bm{1}}]\right] \notag \\
    &- \E_{q(N \mid \gD)} \left[{\textstyle\sum}_{n=1}^{N}\,\E_{q(\rvz_n^\where \mid \gD)}[\KL[q(\rvo_{n,\gD}^{\mathrm{2D}} \!\mid\! \rvz_n^\where, \gD) \!\parallel\! \N{\bm{0}, \bm{1}}]]\right]\ .
\end{align*}
Here, $\gamma$ is a weighting hyperparameter, and ${\mathrm{Geom}}(\rho)$ is a truncated Geometric distribution with support $\{0, 1, \dots, N_{\mathrm{max}}\}$ and success probability $\rho$. We set $\gamma = 7$ and $\rho = 0.999$ during training, thereby encouraging the model to decompose the scenes into as few objects as possible.

\section{Related Work} \label{sec:related_work}

ROOTS is broadly related to recent advances in learning representations for the appearance and geometry of 3D scenes, and more closely related to those that do not require 3D supervision. ROOTS is also inspired by recent works that learn to decompose 2D scenes into object-wise representations.

\textbf{Geometric Deep Learning.}
Learning representations that capture the geometry of 3D scenes has been of growing interest. Recent works have explored integrating voxels \citep{maturana2015voxnet, kar2017learning, tulsiani2017multi, wu2016learning, choy20163d}, meshes \citep{kato2018neural, kanazawa2018learning}, point clouds \citep{qi2017pointnet, achlioptas2018learning}, and many other classical representations into deep learning models to achieve better 3D scene understanding. However, they often require 3D supervision \citep{huang2018cooperative, tulsiani2018factoring, cheng2018geometry, shin20193d, du2018learning} and work on single-object scenes \citep{wu2016learning, yan2016perspective, choy20163d,  kar2017learning, nguyen2019hologan}. By contrast, ROOTS learns to decompose a multi-object scene into object-wise representations without any 3D supervision.

\textbf{Neural Representation of 3D Scenes.}
Recent works \citep{gqn,cgqn,tobin2019geometry,grnn,sitzmann2019deepvoxels,srn,snp,nerf,enr} have explored learning 3D scene representations from 2D images without 3D supervision. While the rendering quality \citep{tobin2019geometry,srn,nerf} and efficiency \citep{enr} have been improved, these methods are not able to decompose the full scene into objects without object-level supervision, and cannot learn object-wise representation and rendering models. We believe these works are complementary to ROOTS and may allow object models to be learned from more realistic scenes. \citet{srn_silot} recently proposed to learn 3D object-centric representations from unlabeled videos. Although their model can infer the 3D position of each object, the object appearance is modeled in 2D. Another line of work \citep{blockgan,liao2019towards} learns object-aware 3D scene representations for generative adversarial networks \citep{gan}. They only support rendering and are unable to infer the object models for a given scene.

\textbf{Object-Oriented Representation of 2D Images.}
There have been prolific advances in unsupervised object-oriented representation learning from fully-observed 2D images. They mainly fall into two categories: detection-based and mixture-based. The detection-based approaches \citep{air,spair,space} first identify object regions and then learn object representations from object patches cropped by the spatial transformer \citep{spatial_transformer}. The mixture-based approaches \citep{nem,monet,iodine,genesis} model the observed image as a pixel-level Gaussian mixture where each component is expected to capture a single object. None of these approaches consider the 3D structure of the scene, let alone the 3D appearance of objects.

\section{Experiments}

In this section, we evaluate the quality of object models learned by ROOTS and demonstrate the benefits they bring in terms of generation quality, generalization ability, and downstream task performance. We also showcase the built-in compositionality and disentanglement properties of ROOTS. We first introduce the data sets and baselines we use, and then show both qualitative and quantitative results.


\textbf{Data Sets.}
Existing data sets in previous work on unsupervised 3D scene representation learning \citep{gqn, tobin2019geometry} either do not contain multi-object scenes or cannot provide object-wise groundtruth information like object positions, and thus cannot serve our purpose. Hence, we created two data sets: the Shapes data set and the Multi-Shepard-Metzler (MSM) data set, using MuJoCo~\citep{mujoco} and Blender~\citep{blender} respectively. Both data sets contain 60K multi-object scenes (50K for training, 5K for validation, and 5K for testing) with complete groundtruth scene specifications. Each scene is rendered as 128$\times$128 color images from 30 random viewpoints. Notice that the scene specifications are for evaluation only and are not used during training. 

We generated three versions of the Shapes data set, containing scenes with 1-3, 2-4, and 3-5 objects respectively. The position, size, shape, and color of the objects are randomized. The MSM data set contains scenes with 2-4 randomly positioned Shepard-Metzler objects. Each object consists of 5 cubes whose positions are generated by a self-avoiding random walk. The color of each cube is independently sampled from a continuous color space, as described in GQN~\citep{gqn}. Since these objects have complex shapes randomly generated per scene, they span a large combinatorial space, and it is unlikely that two different scenes will share a same object. Also, the objects can have severe occlusion with each other, making this data set significantly more challenging than the single-object version considered in GQN.

For evaluation on realistic objects, we also included a publicly available ShapeNet arrangement data set \citep{grnn, cheng2018geometry}. Each scene of this data set consists of 2 ShapeNet~\citep{shapenet2015} objects placed on a table surface, and is rendered from 54 fixed cameras positioned on the upper hemisphere.
Following prior work, we split the data set into a training set of 300 scenes and a test set of 32 scenes containing unseen objects. Because object-wise annotations are not available, we did not perform quantitative evaluation of object-level decomposition on this data set.


\newcommand{\msm}{Multi-Shepard-Metzler}

\textbf{Baselines.} Because there is no previous work that can build 3D object models from multi-object scene images, we use separate baselines to evaluate scene-level representation and object-level decomposition respectively. For scene-level representation and generation quality, we use CGQN~\citep{cgqn} as the baseline model, and refer to it as GQN in the rest of this section to indicate the general GQN framework.
For object-level decomposition, we compare the image segmentation ability embedded in ROOTS with that of IODINE~\citep{iodine}, which focuses on this ability without learning 3D representations.


\begin{figure}[t]
\centering
\includegraphics[width=\linewidth]{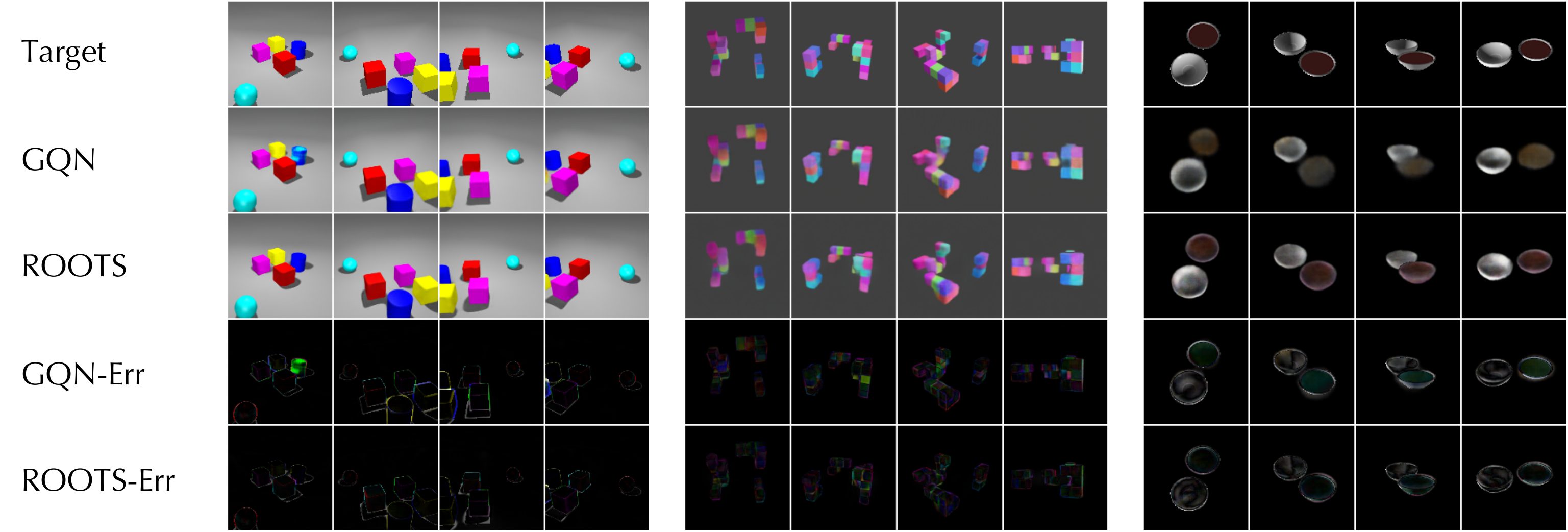}
\caption{Sample generations from three scenes. Columns correspond to query viewpoints. ROOTS gives better generations in regions of occlusion while GQN sometimes misses occluded objects and predicts wrong colors. GQN-Err and ROOTS-Err are difference maps between targets and generations of GQN and ROOTS, respectively.}
\label{generation_vis}
\end{figure}

\subsection{Qualitative Evaluation}
In this section, we qualitatively evaluate the learned object models by showing scene generation and decomposition results and object model visualizations. We also demonstrate the built-in compositionality and disentanglement properties by compositing novel scenes out of the training distribution and visualizing latent traversals, respectively.


\textbf{Scene Generation.} Like GQN, ROOTS is able to generate target observations for a given scene from arbitrary query viewpoints. \Figref{generation_vis} shows a comparison of scene generations using 15 contexts. ROOTS gives better generations in regions of occlusion (especially on the MSM data set), and correctly infers partially observable objects (e.g., the yellow cube in the 4th column). In contrast, GQN tends to miss heavily occluded and partially observable objects, and sometimes predicts wrong colors. As highlighted in the difference maps in \Figref{generation_vis}, on the Shapes data set, GQN sometimes generates inconsistent colors within an object. On the MSM data set, GQN samples may look slightly clearer than those of ROOTS as GQN generates sharper boundaries between the unit cubes. However, the difference map reveals that GQN more frequently draws the objects with wrong colors. On the ShapeNet arrangement data set, GQN samples are more blurry and also with wrong colors. We believe that the object models learned by ROOTS and the object-level modular rendering provide ROOTS with a stronger capacity to represent the appearance of individual objects, leading to its better generation quality.

\begin{figure}[p]
\centering
\includegraphics[width=0.9\linewidth]{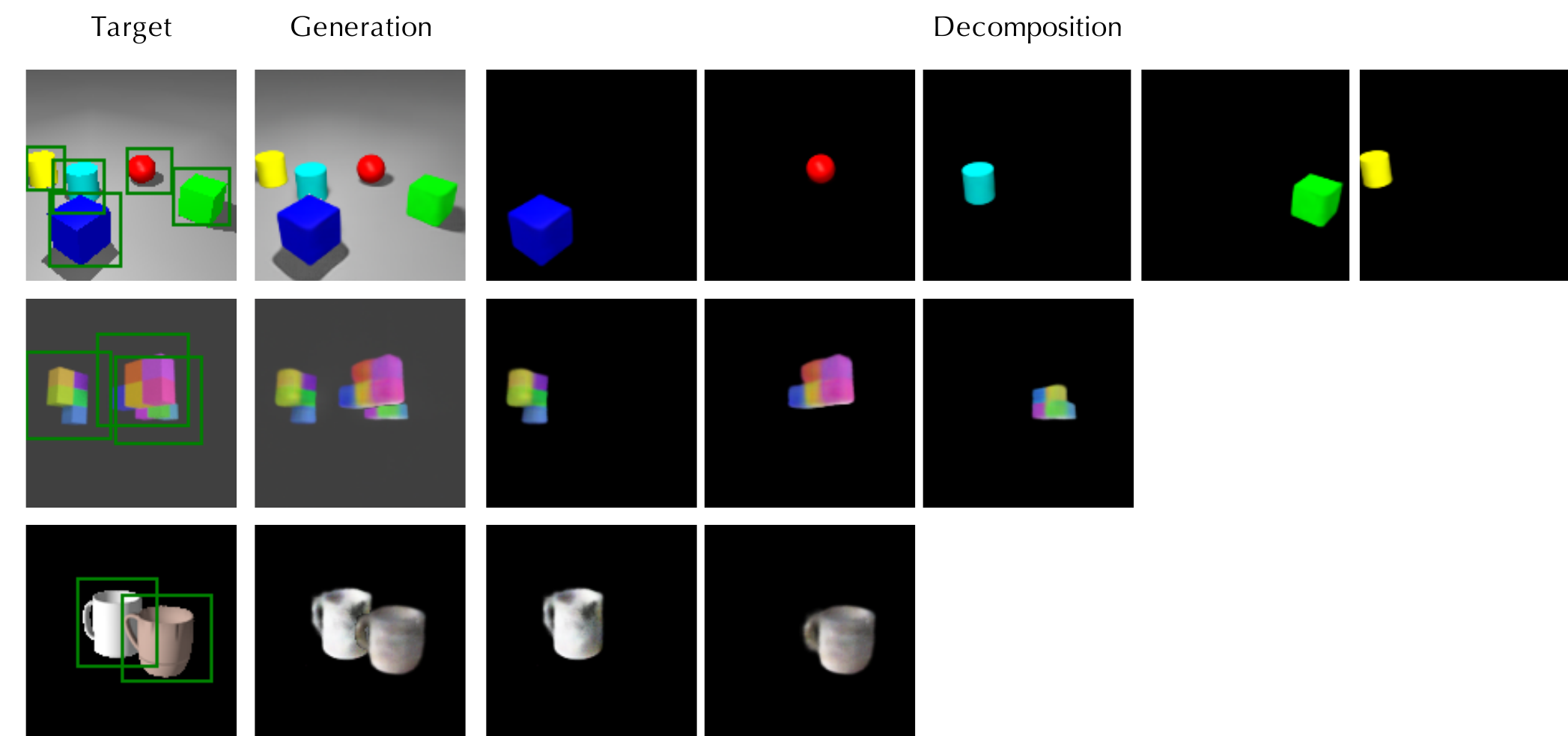}
\caption{The full scene is composited from individual object rendering results. Predicted bounding boxes are drawn on target images.}
\label{gen_decom_vis}
\end{figure}

\begin{figure}[p]
\centering
\includegraphics[width=\linewidth]{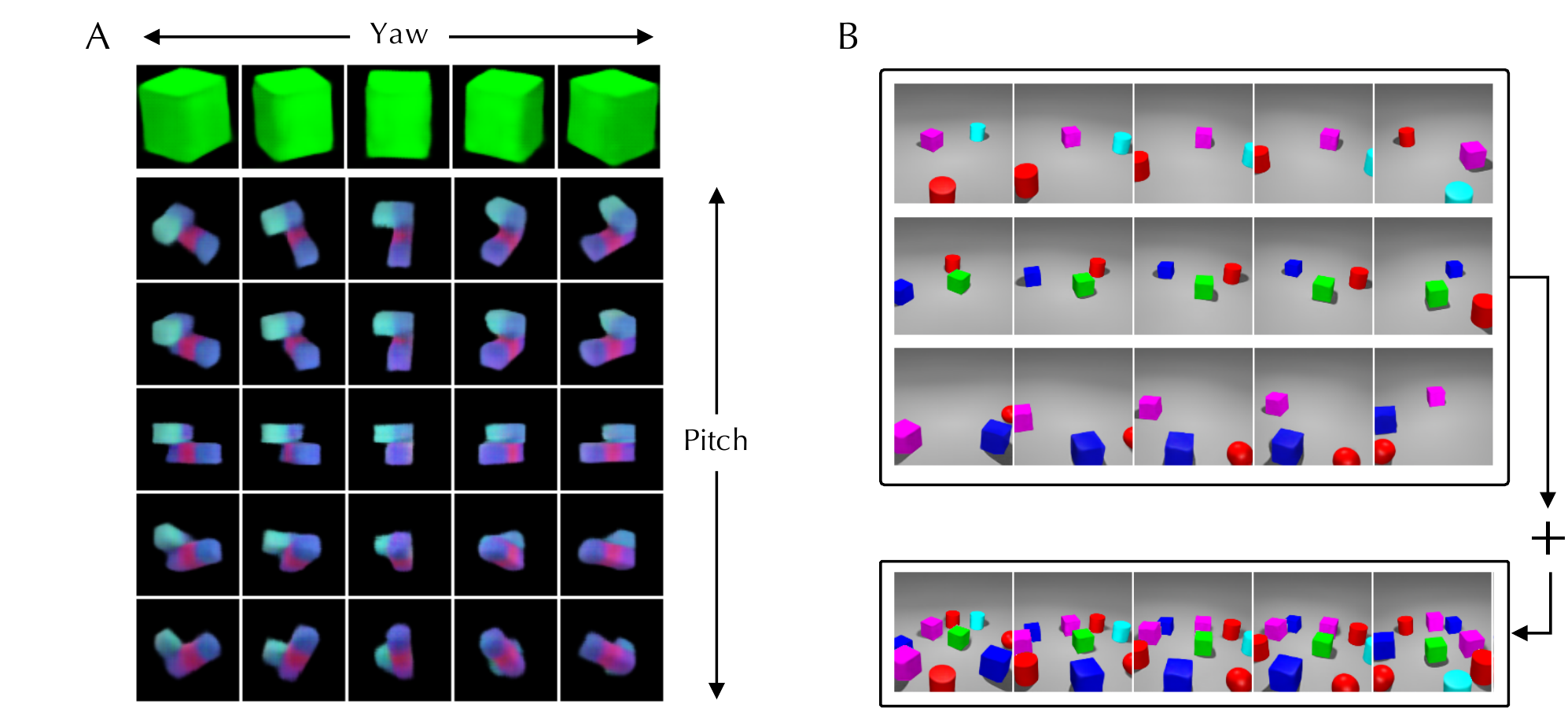}
\caption{{\em (A)} Visualization of learned object models from a set of query viewpoints. {\em (B)} Learned object models are reconfigured into a novel scene. Columns correspond to query viewpoints.}
\label{obj_model_and_compositionality}
\end{figure}

\textbf{Object Models.} We further visualize the learned object models in \Figref{obj_model_and_compositionality}A, by applying the object renderer to $\rvz_n^\what$ and a set of query viewpoints. We also show the scene rendering process in \Figref{gen_decom_vis}, where object rendering results are composited to generate the full scene. As can be seen, from images containing multiple objects with occlusion, ROOTS is able to learn the complete 3D appearance of each object, predict accurate object positions, and correctly handle occlusion. Such object models are not available from GQN because it only learns scene-level representations.

\begin{figure}[t]
\centering
\includegraphics[width=0.9\textwidth]{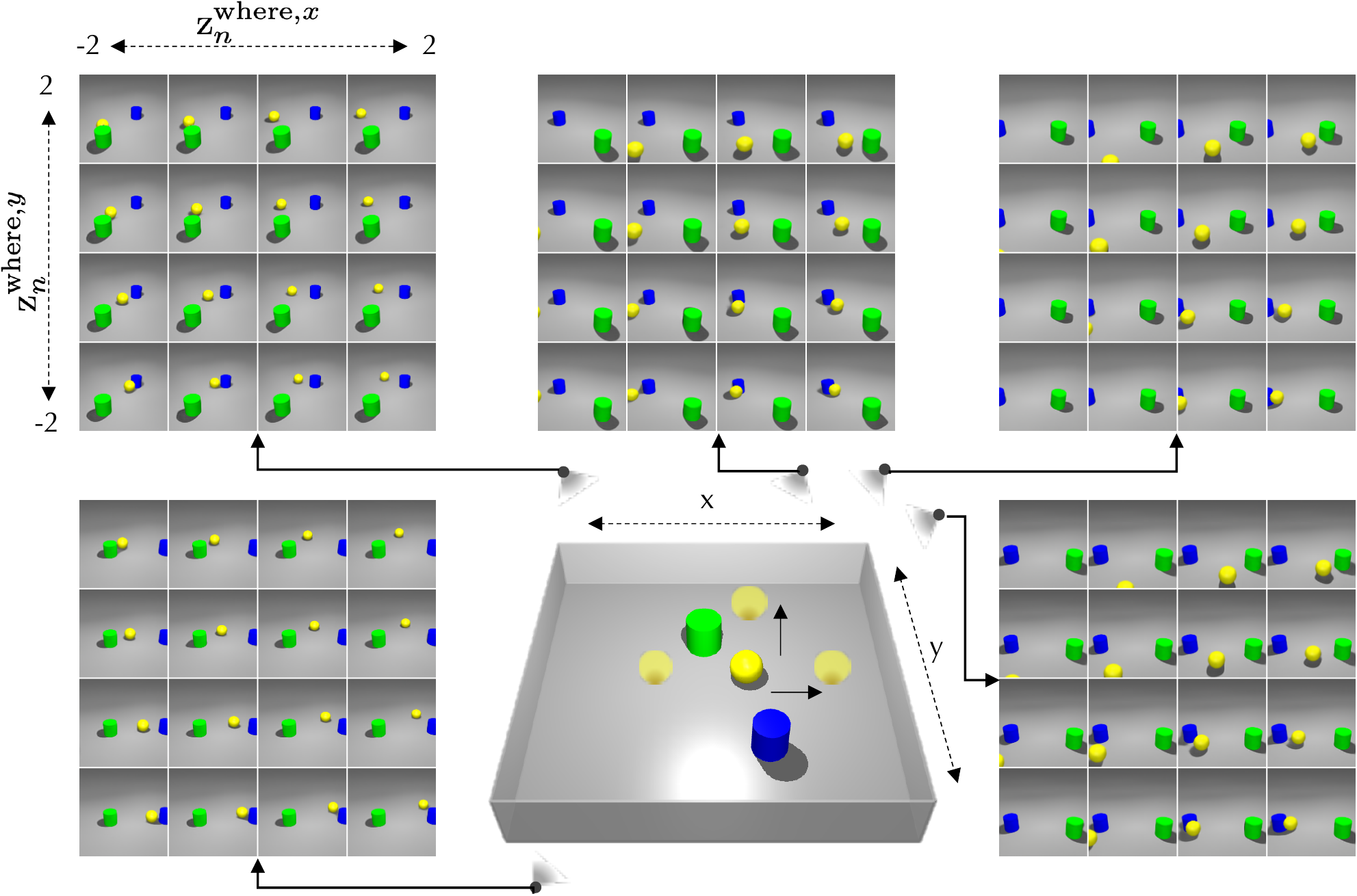}
\caption{Traversal of the position latent $\rz^{\where, x}_n$ and $\rz^{\where, y}_n$ of the yellow ball in the scene. We show generations from five query viewpoints after manipulating the position latent.}
\label{disentangle}
\end{figure}

\textbf{Compositionality.}
Once object models are learned, they can be reconfigured to form novel scenes that are out of the training distribution. As an example, in \Figref{obj_model_and_compositionality}B, we first provide ROOTS with context images from three scenes (top three rows) with 3 objects each, and collect the learned object representations $\{(\bm{r}_n^\att, \rvz_n^\where, \rvz_n^\what)\}$. A new scene with 9 objects can then be composed and rendered from arbitrary query viewpoints.
Rendering results are shown in the bottom row of \Figref{obj_model_and_compositionality}B. We would like to emphasize that the model is trained on scenes with 1-3 objects. Thus, a scene with 9 objects has never been seen during training.

\textbf{Disentanglement.}
Since object position and appearance are disentangled in the learned object models, by manipulating the position latent, we are able to move objects around without changing other factors like object appearance. 
In \Figref{disentangle}, we visualize traversals of $\rz^{\where, x}_n$ and $\rz^{\where, y}_n$ of the yellow ball through generations from 5 query viewpoints. It can be seen that the change of one coordinate does not affect the other. In addition, the appearance of the yellow ball remains complete and clean during the traversal. Other untouched components (the green cylinder, the blue cylinder, and the background) remain unchanged. Moreover, we also notice some desired rendering effects. For example, the size of the yellow ball becomes smaller as it moves further away from the camera.

\subsection{Quantitative Evaluation}
In this section, we report quantitative results on scene generation and decomposition, which reflect the quality of the learned object models. We also highlight the benefit of learning object models in two downstream tasks.

\textbf{Scene Generation.}
To compare the generation quality of ROOTS and GQN, in Table~\ref{ELBO_MSE_1} and Table~\ref{ELBO_MSE_2}, we report negative log-likelihood (NLL) and mean squared error (MSE) on the test sets. We provide 15 context observations for both models, and use 100 samples to approximate NLL. Similar to previous works \citep{cgqn,babaeizadeh2018stochastic}, we report the minimum MSE over 100 samples from the learned conditional prior. This measures the ability of a conditional generative model to capture the true outcome within its conditional prior of all possible outcomes. ROOTS outperforms GQN on both metrics, showing that learning object models also contributes to better generation quality.
\begin{table}[t]
\begin{center}
\begin{adjustbox}{max width=\textwidth}
\begin{tabular}{ccccccc}
\toprule
\multicolumn{1}{c}{Data Set}  &\multicolumn{2}{c}{1-3 Shapes}  &\multicolumn{2}{c}{2-4 Shapes} &\multicolumn{2}{c}{3-5 Shapes}\\
\cmidrule(lr){1-1} \cmidrule(lr){2-3} \cmidrule(lr){4-5} \cmidrule(lr){6-7}
\multicolumn{1}{c}{Metrics} &\multicolumn{1}{c}{NLL$\downarrow$}  &\multicolumn{1}{c}{MSE$\downarrow$} &\multicolumn{1}{c}{NLL$\downarrow$}  &\multicolumn{1}{c}{MSE$\downarrow$} &\multicolumn{1}{c}{NLL$\downarrow$}  &\multicolumn{1}{c}{MSE$\downarrow$}\\
\midrule
ROOTS             &-207595.81 &30.60 &-206611.07 &42.41 &-205608.07 &54.45 \\ 
GQN            &-206760.87 &40.62 &-205604.74 &54.49 &-204918.39 &62.73 \\ 
\bottomrule
\end{tabular}
\end{adjustbox}
\end{center}
\caption{Quantitative evaluation of scene generation on the Shapes data sets.}
\label{ELBO_MSE_1}
\end{table}

\begin{table}[t]
\begin{center}
\begin{adjustbox}{max width=\textwidth}
\begin{tabular}{ccccc}
\toprule
\multicolumn{1}{c}{Data Set}  &\multicolumn{2}{c}{Multi-Shepard-Metzler} &\multicolumn{2}{c}{ShapeNet Arrangement}\\
\cmidrule(lr){1-1} \cmidrule(lr){2-3} \cmidrule(lr){4-5}
\multicolumn{1}{c}{Metrics} &\multicolumn{1}{c}{NLL$\downarrow$}  &\multicolumn{1}{c}{MSE$\downarrow$}
&\multicolumn{1}{c}{NLL$\downarrow$}  &\multicolumn{1}{c}{MSE$\downarrow$}\\
\midrule
ROOTS             &-206627.56 &42.22 &-192414.85 &212.77\\ 
GQN            &-206294.22 &46.22 &-185010.31 &301.62\\ 
\bottomrule
\end{tabular}
\end{adjustbox}
\end{center}
\caption{Quantitative evaluation of scene generation on the Multi-Shepard-Metzler data set and the ShapeNet arrangement data set.}
\label{ELBO_MSE_2}
\end{table}

\begin{figure}[t]
	\centering
	\begin{tikzpicture}[scale=0.64]
	\begin{groupplot}[group style={group size=2 by 1,horizontal sep = 100pt,vertical sep = 0pt}]
	\centering
	\nextgroupplot[
	width=7cm,height=7cm,
	scale only axis=true,
	xlabel={Number of Contexts},
	ylabel={Average Precision},
	xmin=5, xmax=25,
	ymin=0.6, ymax=1.0,
	xtick={5,10,15,20,25},
	ytick={0.6,0.7,0.8,0.9,1.0},
	xmajorgrids=true,
	ymajorgrids=true,
	grid style=dashed,
	legend pos=south east,
	legend style={font={\fontsize{14 pt}{14 pt}\selectfont}},
	label style={font={\fontsize{14 pt}{14 pt}\selectfont}},
	tick label style={font={\fontsize{12 pt}{12 pt}\selectfont}},
	]
	\addplot[
	color=violet,
	mark=diamond,
	line width=1.2pt,
	mark size=4.2pt,
	]
	coordinates {(5,0.85765004)
		 (10,0.9165649)
		(15,0.93134874)
		(20,0.9408515)
		(25,0.9404509)
	};
	\addplot[
	color=green,
	mark=triangle,
	line width=1.2pt,
	mark size=4.2pt,
	]
	coordinates { (5,0.75795037)
		(10,0.8511123)
		(15,0.8815975)
		(20,0.8943312)
		(25,0.89938116)
	};
	\addplot[
	color=red,
	mark=o,
	line width=1.2pt,
	mark size=3.6pt,
	]
	coordinates {(5,0.6640706)
		(10,0.77967435)
		(15,0.81877536)
		(20,0.84163123)
		(25,0.85022944)
	};
	\addplot[
	color=blue,
	mark=star,
	line width=1.2pt,
	mark size=3.6pt,
	]
	coordinates {(5,0.824627)
		(10,0.8687641)
		(15,0.8808238)
		(20,0.8852957)
		(25,0.8856092)
	};
	\legend{1-3 Shapes,2-4 Shapes,3-5 Shapes, MSM}
	\nextgroupplot[
	width=7cm,height=7cm,
	scale only axis=true,
	xlabel={Number of Contexts},
	ylabel={Counting Accuracy},
	xmin=5, xmax=25,
	ymin=0.8, ymax=1.0,
	xtick={5,10,15,20,25},
	ytick={0.8,0.85,0.9,0.95,1.},
	xmajorgrids=true,
	ymajorgrids=true,
	grid style=dashed,
	legend pos=south east,
	legend cell align={left},
	legend style={font={\fontsize{14 pt}{14 pt}\selectfont}},
	label style={font={\fontsize{14 pt}{14 pt}\selectfont}},
	tick label style={font={\fontsize{12 pt}{12 pt}\selectfont}},
	]
	\addplot[
	color=violet,
	mark=diamond,
	line width=1.2pt,
	mark size=4.2pt,
	]
	coordinates {(5,0.9602)
		(10,0.9836)
		(15,0.9830)
		(20,0.9876)
		(25,0.9870)
	};
	\addplot[
	color=green,
	mark=triangle,
	line width=1.2pt,
	mark size=4.2pt,
	]
	coordinates { 
		(5,0.9244)
		(10,0.9592)
		(15,0.9720)
		(20,0.9750)
		(25,0.9734)
	};
	\addplot[
	color=red,
	mark=o,
	line width=1.2pt,
	mark size=3.6pt,
	]
	coordinates {
		(5,0.8756)
		(10,0.9488)
		(15,0.9596)
		(20,0.9638)
		(25,0.9656)
	};
	\addplot[
	color=blue,
	mark=star,
	line width=1.2pt,
	mark size=3.6pt,
	]
	coordinates {(5,0.9584)
		(10,0.9762)
		(15,0.9786)
		(20,0.9776)
		(25,0.9784)
	};
	\end{groupplot}    
	\end{tikzpicture}
	\caption{Average precision and counting accuracy.}
	\label{ap_acc}
\end{figure}
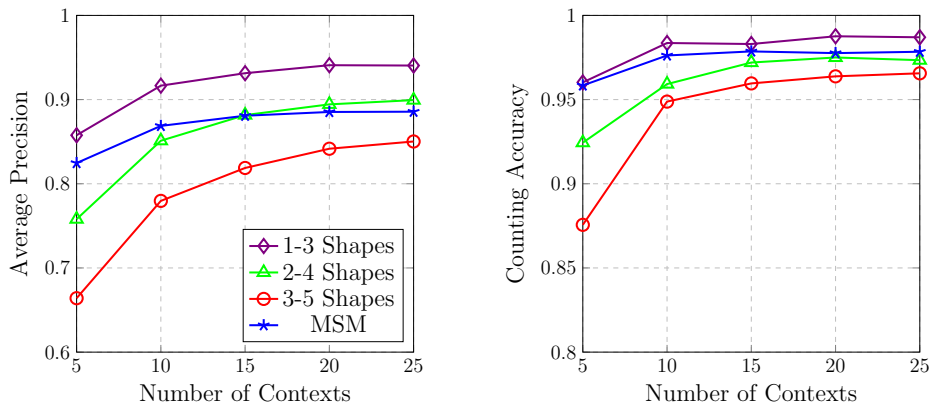

\textbf{Object Models.}
To evaluate the quality of learned object models, we report object counting accuracy and an adapted version of Average Precision (AP, \citealt{everingham2010pascal}) in \Figref{ap_acc}.
AP measures the object localization ability. To compute AP, 
we set some thresholds $t_i$ on the 3D distance between the predicted $\rvz_n^\where$ and the groundtruth object center position. If the distance is within the threshold, the prediction is considered a true positive. Clearly, a smaller threshold requires the model to locate objects more accurately. We set three thresholds: $1/4$, $2/4$, and $3/4$ of the average object size. For each threshold $t_i$, we obtain the area under the precision-recall curve as $\text{AP}(t_i)$. The final AP is averaged over the three thresholds: ${\text{AP} = \sum_{i=1}^3 \text{AP}(t_i) / 3}$. We vary the number of contexts provided, and compute counting accuracy and AP using the predicted $N$ and $\rvz_n^\where$ that achieve the minimum MSE over 10 samples from the conditional prior. As shown in \Figref{ap_acc}, both counting accuracy and AP increase as the number of context observations becomes larger. This indicates that ROOTS can effectively accumulate information from the given contexts.

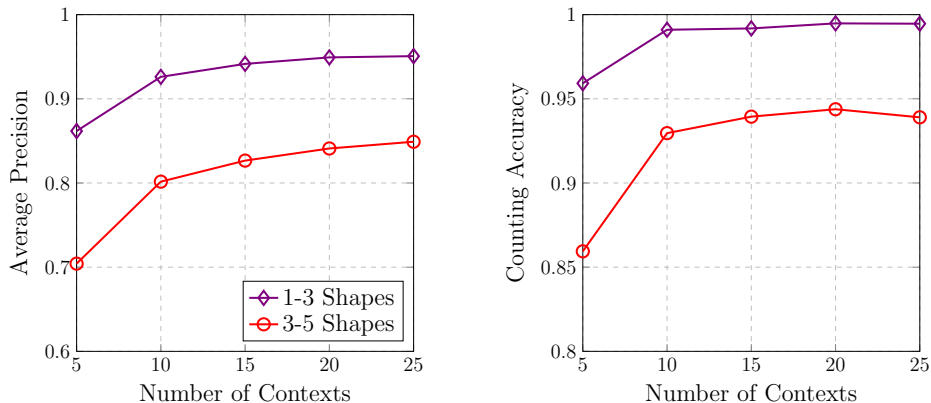
\begin{figure}[t]
	\centering
	\begin{tikzpicture}[scale=0.64]
	\begin{groupplot}[group style={group size=2 by 1,horizontal sep = 100pt,vertical sep = 0pt}]
	\centering
	\nextgroupplot[
	width=7cm,height=7cm,
	scale only axis=true,
	xlabel={Number of Contexts},
	ylabel={Average Precision},
	xmin=5, xmax=25,
	ymin=0.6, ymax=1.0,
	xtick={5,10,15,20,25},
	ytick={0.6,0.7,0.8,0.9,1.0},
	xmajorgrids=true,
	ymajorgrids=true,
	grid style=dashed,
	legend pos=south east,
	legend style={font={\fontsize{14 pt}{14 pt}\selectfont}},
	label style={font={\fontsize{14 pt}{14 pt}\selectfont}},
	tick label style={font={\fontsize{12 pt}{12 pt}\selectfont}},
	]
	\addplot[
	color=violet,
	mark=diamond,
	line width=1.2pt,
	mark size=4.2pt,
	]
	coordinates {(5,0.8617)
		(10,0.9261)
		(15,0.9415)
		(20,0.9492)
		(25,0.9507)
	};
	\addplot[
	color=red,
	mark=o,
	line width=1.2pt,
	mark size=3.6pt,
	]
	coordinates { (5,0.7042)
		(10,0.8016)
		(15,0.8266)
		(20,0.8410)
		(25,0.8490)
	};
	\legend{1-3 Shapes,3-5 Shapes}
	\nextgroupplot[
	width=7cm,height=7cm,
	scale only axis=true,
	xlabel={Number of Contexts},
	ylabel={Counting Accuracy},
	xmin=5, xmax=25,
	ymin=0.8, ymax=1.0,
	xtick={5,10,15,20,25},
	ytick={0.8,0.85,0.9,0.95,1.},
	xmajorgrids=true,
	ymajorgrids=true,
	grid style=dashed,
	legend pos=south east,
	legend cell align={left},
	legend style={font={\fontsize{14 pt}{14 pt}\selectfont}},
	label style={font={\fontsize{14 pt}{14 pt}\selectfont}},
	tick label style={font={\fontsize{12 pt}{12 pt}\selectfont}},
	]
	\addplot[
	color=violet,
	mark=diamond,
	line width=1.2pt,
	mark size=4.2pt,
	]
	coordinates {(5,0.9592)
		(10,0.9910)
		(15,0.9918)
		(20,0.9948)
		(25,0.9946)
	};
	\addplot[
	color=red,
	mark=o,
	line width=1.2pt,
	mark size=3.6pt,
	]
	coordinates { 
		(5,0.8594)
		(10,0.9296)
		(15,0.9394)
		(20,0.9438)
		(25,0.9390)
	};
	\end{groupplot}    
	\end{tikzpicture}
	\caption{Generalization performance of average precision and counting accuracy. ROOTS is trained on the 2-4 Shapes data set.}
	\label{ap_acc_generalize}
\end{figure}

\textbf{Segmentation of 2D Observations.}
The rendering process of ROOTS implicitly segments 2D observations under query viewpoints. The segmentation performance reflects the quality of learned 3D object appearance. Since GQN cannot provide such segmentation, we compare ROOTS with IODINE~\citep{iodine} in terms of the Adjusted Rand Index (ARI, \citealt{rand1971,hubert1985comparing}) on the Shapes data sets (IODINE completely failed on the MSM data set---it tends to split one object into multiple slots based on color similarity, as we show in Appendix \ref{app:iodine}). We train IODINE on all the images available in the training set, using the official implementation. At test time, ROOTS is given 15 random contexts for each scene and performs segmentation for an unseen query viewpoint. ROOTS does not have access to the target image under the query viewpoint. In contrast, IODINE directly takes the target image as input. 
Results in Table~\ref{ari} show that ROOTS outperforms IODINE on both foreground segmentation (ARI-NoBg) and full image segmentation (ARI). We would like to emphasize that IODINE specializes in 2D scene segmentation, whereas ROOTS obtains its 2D segmentation ability as a by-product of learning 3D object models.
\begin{table}[t]
	\centering
	\begin{adjustbox}{max width=\textwidth}
		\begin{tabular}{ccccccccc}
			\toprule
			\multicolumn{1}{c}{Data Set} &\multicolumn{2}{c}{1-3 Shapes} &\multicolumn{2}{c}{2-4 Shapes} &\multicolumn{2}{c}{3-5 Shapes} &\multicolumn{2}{c}{Multi-Shepard-Metzler} \\
			\cmidrule(lr){1-1} \cmidrule(lr){2-3} \cmidrule(lr){4-5}  \cmidrule(lr){6-7} \cmidrule(lr){8-9}
			\multicolumn{1}{c}{Metrics} &\multicolumn{1}{c}{ARI$\uparrow$} &\multicolumn{1}{c}{ARI-NoBg$\uparrow$} &\multicolumn{1}{c}{ARI$\uparrow$} &\multicolumn{1}{c}{ARI-NoBg$\uparrow$} &\multicolumn{1}{c}{ARI$\uparrow$} &\multicolumn{1}{c}{ARI-NoBg$\uparrow$}  &\multicolumn{1}{c}{ARI$\uparrow$} &\multicolumn{1}{c}{ARI-NoBg$\uparrow$} \\
			\midrule
			\multicolumn{1}{c}{ROOTS} &0.9477 &0.9942 &0.9482  &0.9947 &0.9490  &0.9930 & 0.9303 &0.9608 \\
			\multicolumn{1}{c}{IODINE} &0.8217 &0.8685 &0.8348 &0.9854 &0.8422  &0.9580 & Failed & Failed \\
			\bottomrule
		\end{tabular}
	\end{adjustbox}
	\caption{Quantitative evaluation of 2D segmentation.}
	\label{ari}
\end{table}

\textbf{Generalization.}
To evaluate the generalization ability, we first train ROOTS and GQN on the Shapes data set with 2-4 objects, and then test on the Shapes data sets with 1-3 objects and 3-5 objects respectively. As shown in Table~\ref{nll_mse_generalization}, ROOTS achieves better NLL and MSE in both interpolation and extrapolation settings. We further report AP and counting accuracy for ROOTS when generalizing to the above two data sets. As shown in \Figref{ap_acc_generalize}, ROOTS generalizes well to scenes with 1-3 objects, and performs reasonably when given more context observations on scenes with 3-5 objects.

\begin{table}[t]
        \centering
        \begin{adjustbox}{max width=\textwidth}
        \begin{tabular}{ccccc}
	        \toprule
	        \multicolumn{1}{c}{Training Set} &\multicolumn{4}{c}{2-4 Shapes} \\
	        \cmidrule(lr){1-1} \cmidrule(lr){2-5}
	        \multicolumn{1}{c}{Test Set} &\multicolumn{2}{c}{1-3 Shapes} &\multicolumn{2}{c}{3-5 Shapes} \\ \cmidrule(lr){1-1} \cmidrule(lr){2-3} \cmidrule(lr){4-5}
	        \multicolumn{1}{c}{Metrics} &\multicolumn{1}{c}{NLL$\downarrow$} &\multicolumn{1}{c}{MSE$\downarrow$} &\multicolumn{1}{c}{NLL$\downarrow$}  &\multicolumn{1}{c}{MSE$\downarrow$} \\
	        \midrule
	        \multicolumn{1}{c}{ROOTS} &-208122.58 &24.27 &-204480.37  &67.98 \\
	        \multicolumn{1}{c}{GQN} &-207616.49 &30.35 &-202922.03 &86.68 \\
	        \bottomrule
        \end{tabular}
        \end{adjustbox}
        \caption{Quantitative evaluation of generalization ability.}
        \label{nll_mse_generalization}
\end{table}

\begin{table}[t]
\vskip 0.2in
        \centering
        \begin{adjustbox}{max width=0.97\textwidth}
        \begin{tabular}{cccc}
            \toprule
			\multirow{2}{*}{Tasks} &\multicolumn{2}{c}{Retrieve Object} &\multirow{2}{*}{Find Pair}\\ 
			\cmidrule(lr){2-3}
			
			& 3D Version  & 2D Version \\
			
			\midrule
			ROOTS     &90.38\%     & 93.71\%     & 84.70\%\\
			GQN       &81.31\%     & 84.18\%     & 12.48\%\\
			\bottomrule
	        
        \end{tabular}
        \end{adjustbox}
        \caption{Testing accuracies on downstream tasks.}
        \label{downstream}
\end{table}


\textbf{Downstream 3D Reasoning Tasks.} The 3D object models can facilitate object-wise 3D reasoning. We demonstrate this in two downstream tasks on the Shapes data set with 3-5 objects. 
\textbf{\textit{Retrieve Object.}} The goal of this task is to retrieve the object that lies closest to a given position $\bm{p}$. We consider both 3D and 2D versions of the task. In 3D version, we set $\bm{p}$ as the origin of the 3D space, whereas in 2D version, $\bm{p}$ is the center point of the target image from viewpoint $\rvv_q$. We treat this task as a classification problem, where the input is the learned representation  (along with $\rvv_q$ in 2D version), and the output is the label of the desired object. Here, the label is an integer assigned to each object based on its shape and color. 
We compare ROOTS with the GQN baseline, and report testing accuracies in Table~\ref{downstream}. ROOTS outperforms GQN, demonstrating the effectiveness of the learned object models in spatial reasoning.
\textbf{\textit{Find Pair.}} In this task, the goal is to find two objects that have the smallest pair-wise distance in 3D space. Again, we treat this as a classification task, where the target label is the sum of labels of the two desired objects. The testing accuracies are reported in Table~\ref{downstream}. Clearly, this task requires pair-wise relational reasoning. The object models learned by ROOTS naturally allows extraction of pair-wise relations.
In contrast, the scene-level representation of GQN without object-wise factorization leads to incompetence in relational reasoning. 

\subsection{Ablation Study}

Our ablation study shows that the components of ROOTS are necessary for obtaining object models. In particular, we tried the following alternative design choices.

\textbf{{ROOTS Encoder.}} One may think that
${\rvz_n^\what}$ can be directly inferred from scene-level contexts without object-attention grouping. Thus, we tried inferring ${\rvz_n^\what}$ from GVFM along with ${\rvz_n^\where}$. The model, however, failed to decompose scenes into objects and hence was not trainable.

\textbf{{ROOTS Decoder.}} One may also think that the object-specific image layer $\hat{\rvx}_{n,q}$ can be directly generated from the 3D object model $\rvz_n^\mathrm{3D}$ without having the intermediate 2D representation $\rvo_{n,q}^\mathrm{2D}$. This model was also not trainable as it could not use the object positions effectively.  




\section{Conclusion}

We proposed ROOTS, a probabilistic generative model for unsupervised learning of 3D object models from partial observations of multi-object 3D scenes. The learned object models capture the complete 3D appearance of individual objects, yielding better generation quality of the full scene. They also improve generalization ability and allow out-of-distribution scenes to be easily generated. Moreover, in downstream 3D reasoning tasks, ROOTS shows superior performance compared to the baseline model. Interesting future directions would be to learn the knowledge of the 3D world in a sequential manner similarly as we humans keep updating our knowledge of the world.


\acks{We would like to acknowledge support for this project from Kakao Brain and Center for Super Intelligence (CSI). We would like to thank Jindong Jiang, Skand Vishwanath Peri, and Yi-Fu Wu for helpful discussion.}

\newpage
\appendix



\section{Generation Samples}\label{generation_supp}
\begin{figure}[h]
\vskip 0.2in
\centering
\includegraphics[width=0.8\textwidth]{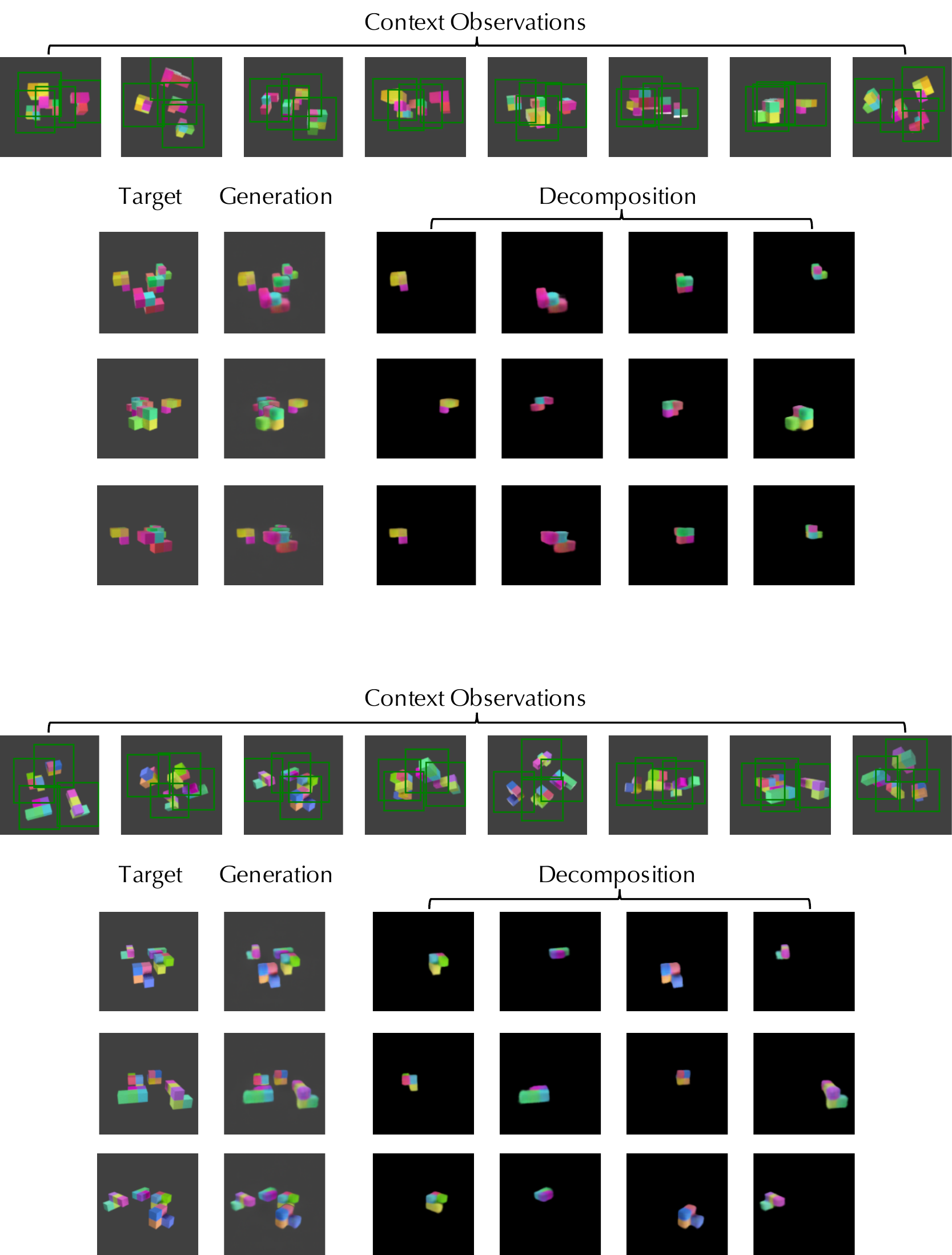}
\caption{Generation samples from the Multi-Shepard-Metzler data set.}
\label{gen_supp1}
\end{figure}

\begin{figure}[h]
\centering
\includegraphics[width=0.8\textwidth]{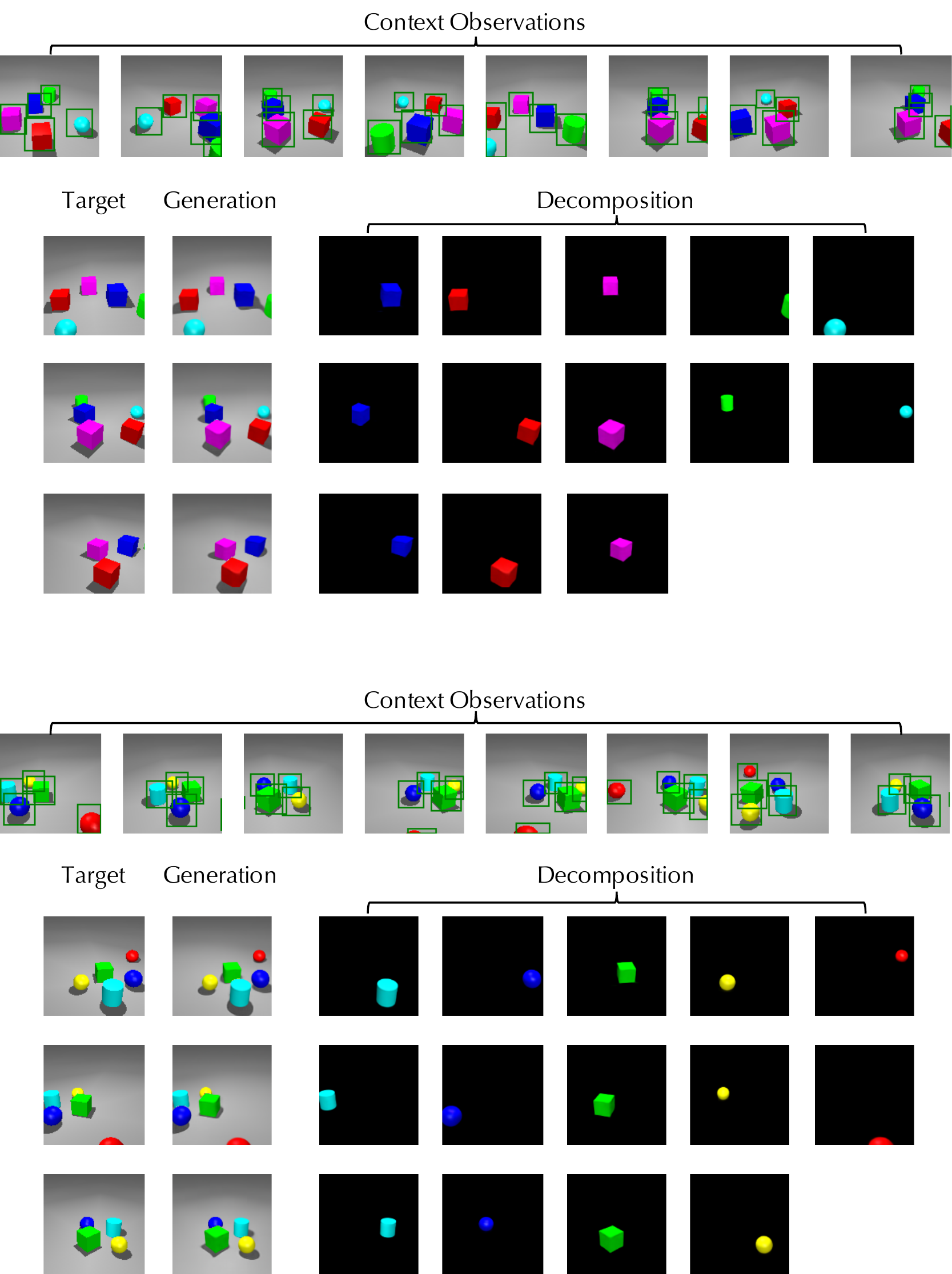}
\caption{Generation samples from the 3-5 Shapes data set.}
\label{gen_supp2}
\end{figure}

We provide more generation samples in this section. For each scene in \Figref{gen_supp1}, we show 8 sampled context images in the top row, superimposed with predicted bounding boxes. We also show generations from three query viewpoints, together with the decomposed object-wise rendering results. Similar visualizations for two scenes from the 3-5 Shapes data set are provided in \Figref{gen_supp2}.

\newpage

\newpage

\section{Object Models}
\begin{figure}[h]
\vskip 0.2in
\centering
\includegraphics[width=.6\textwidth]{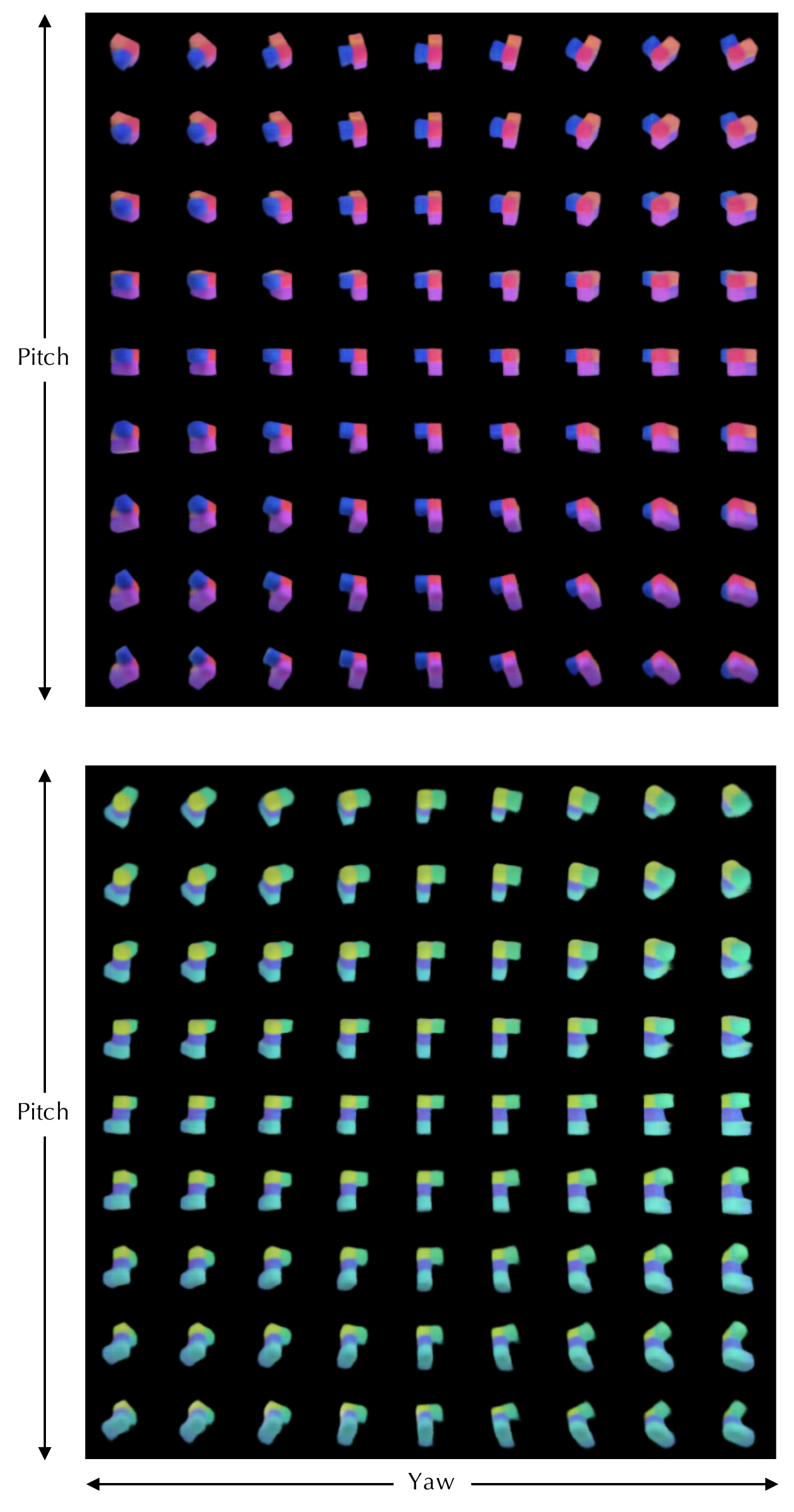}
\caption{Visualization of learned object models from a set of query viewpoints.}
\label{all_views}
\end{figure}

In this section, we provide two more samples of the learned object models. As shown in \Figref{all_views}, each object model inferred from a multi-object scene can generate complete object appearance given different query viewpoints.

\newpage

\section{Summary of ROOTS Encoder and Decoder} \label{app:code}

\begin{algorithm}[h]
\caption{ROOTS Encoder}
\label{algo:enc}
\begin{algorithmic}[1]
    \REQUIRE contexts $\gC = \{(\rvx_c, \rvv_c)\}_{c=1}^{K}$, \  partition resolutions $N_x, N_y, N_z$
    \ENSURE $[J] = \{1, 2, \dots, J\}$ for any natural number $J$
    \STATE Obtain Geometric Volume Feature Map $\bm{r} = f_\mathrm{ctx\_enc}(\gC)$
    \FOR {\textbf{each} $(i,j,k) \in [N_x] \times [N_y] \times [N_z]$ \textbf{parallel}}
        \STATE Infer object presence and position in 3D world coordinates: $\rz_{ijk}^\pres, \rvz_{ijk}^\where \sim f_{\pres, \where}(\bm{r})$
    \ENDFOR
    \STATE Obtain the number of objects $N = {\textstyle\sum}_{ijk}\,\rz_{ijk}^\pres$
    \STATE Map each $(i,j,k)$ with $\rz_{ijk}^\pres = 1$ to an object index $n \in [N]$
    \FOR {\textbf{each} object $n \in [N]$ \textbf{parallel}}
        \FOR {\textbf{each} context $(\rvx_c, \rvv_c) \in \gC$ \textbf{parallel}}
            \STATE Infer 2D object location $\rvo_{n,c}^\where$ using Attention-by-Perspective-Projection
            \STATE Crop 2D object patch $\rvx_{n,c}^\att$ from $\rvx_c$ using $\rvo_{n,c}^\where$
        \ENDFOR
        \STATE Obtain object context ${\gC}_n = \{(\rvx_{n,c}^\att, \rvv_c, \rvo_{n,c}^\where)\}_{c=1}^{K}$
        \STATE Infer 3D object appearance $\rvz_n^\what \sim \GQN_\enc({\gC}_n)$
        \STATE Build object model $\rvz_n^{\mathrm{3D}} = (\rvz_n^\where, \rvz_n^\what)$
    \ENDFOR
    \STATE \textbf{return} object models $\{\rvz_n^{\mathrm{3D}}\}_{n=1}^{N}$, \  object contexts $\{\gC_n\}_{n=1}^{N}$
\end{algorithmic}
\end{algorithm}

\begin{algorithm}[h]
\caption{ROOTS Decoder}
\label{algo:dec}
\begin{algorithmic}[1]
    \REQUIRE object models $\{\rvz_n^{\mathrm{3D}}\}_{n=1}^{N}$, \  object contexts $\{\gC_n\}_{n=1}^{N}$, \  query viewpoints $\rvv_\gQ = \{\rvv_q\}_{q=1}^{M}$
    \ENSURE $[J] = \{1, 2, \dots, J\}$ for any natural number $J$
    \FOR {\textbf{each} query viewpoint $\rvv_q \in \rvv_\gQ$ \textbf{parallel}}
        \FOR {\textbf{each} object $n \in [N]$ \textbf{parallel}}
            \STATE Obtain object context encoding $\bm{r}_n^\att = f_\mathrm{obj\_ctx\_enc}(\gC_n)$
            \STATE Decode 3D appearance into 2D image patch:\\\quad\quad\quad$\rvo_{n,q}^\what = \GQN_\dec(\texttt{concat}[\rvz_n^\what, \bm{r}_n^\att], \rvv_q)$
            \STATE Infer 2D object location $\rvo_{n,q}^\where$ using Attention-by-Perspective-Projection
            \STATE Obtain image layer $\hat{\rvx}_{n,q}$ and transparency map $\bm{\alpha}_{n,q}$ from $\rvo_{n,q}^\what$ and $\rvo_{n,q}^\where$
        \ENDFOR
        \STATE Composite the full image $\hat{\rvx}_{q} = {\textstyle\sum}_{n=1}^{N}\,\bm{\alpha}_{n,q} \odot \hat{\rvx}_{n,q}$
    \ENDFOR
    \STATE \textbf{return} generations $\{\hat{\rvx}_{q}\}_{q=1}^{M}$
\end{algorithmic}
\end{algorithm}

\newpage

\section{Perspective Projection} \label{app:proj}

Following GQN \citep{gqn}, we parameterize the viewpoint $\rvv$ as a tuple $(\rvw, \ry, \rp)$, where $\rvw \in \R^3$ is the position of the camera in world coordinates, and $\ry \in \R$ and $\rp \in \R$ are its yaw and pitch respectively. We also assume access to the intrinsic camera parameters, including focal length $\rf \in \R$ and sensor size, that are the same across all scenes. $\AVP_\pos$ converts the center position of an object $n$ from world coordinates $\rvz_n^\where \in \R^3$ to image coordinates $[\rvo_n^\ctr, \ro_n^\depth]^\top \in \R^3$ as follows:
\begin{equation*}
    [\ra, \rb, \rd]^\top = \rmR_{\ry, \rp} (\rvz_n^\where - \rvw)\ , \quad \rvo_n^\ctr = \mathrm{normalize}([\rf \ra / \rd, \rf \rb / \rd]^\top)\ , \quad \ro_n^\depth = \rd\ .
\end{equation*}
Here, $\rmR_{\ry, \rp}$ is a $3 \times 3$ rotation matrix computed from the camera yaw and pitch, $[\ra, \rb, \rd]^\top$ represents the center position of the object in camera coordinates, and $[\ra, \rb]^\top$ is further normalized into image coordinates $\rvo_n^\ctr$, using the focal length and sensor size, so that the upper-left corner of the image corresponds to $[-1, -1]^\top$ and the lower-right corner corresponds to $[1, 1]^\top$.

\section{Transparency Map} \label{app:alpha}

The transparency map $\bm{\alpha}_{n,q}$ ensures that occlusion among objects is properly handled for a query viewpoint $\rvv_q$. Ideally, $\bm{\alpha}_{n,q}(i,j) = 1$ if, when viewed from $\rvv_q$, the pixel $(i,j)$ is contained in object $n$ and is not occluded by any other object, and $\bm{\alpha}_{n,q}(i,j) = 0$ otherwise. On the other hand, the object mask $\hat{\rvm}_{n,q}$ is expected to capture the non-occluded full object, that is, $\hat{\rvm}_{n,q}(i,j) = 1$ if the pixel $(i,j)$ is contained in object $n$ when viewed from $\rvv_q$, regardless of whether it is occluded or not. Therefore, we compute $\bm{\alpha}_{n,q}$ by masking out occluded pixels from $\hat{\rvm}_{n,q}$:
\begin{equation*}
    \bm{\alpha}_{n,q} = \bm{w}_{n,q} \odot \hat{\rvm}_{n,q}\ ,
\end{equation*}
where $\odot$ is pixel-wise multiplication, and $\bm{w}_{n,q}(i,j) = 1$ if object $n$ is the closest one to the camera among all objects that contain the pixel $(i,j)$. In actual implementation, $\bm{\alpha}_{n,q}$, $\hat{\rvm}_{n,q}$, and $\bm{w}_{n,q}$ are not strictly binary, and we obtain the value of $\bm{w}_{n,q}$ at each pixel $(i,j)$ by the masked softmax over negative depth values:
\begin{equation*}
    \bm{w}_{n,q}(i,j) = \frac{\hat{\rvm}_{n,q}(i,j) \exp{(-\ro_{n,q}^\depth)}}{{\textstyle\sum}_{n=1}^{N}\,\hat{\rvm}_{n,q}(i,j) \exp{(-\ro_{n,q}^\depth)}}\ .
\end{equation*}

\section{Data Set Details}

In this section, we provide details of the two data sets we created.

\textbf{Shapes.}
There are 3 types of objects: cube, sphere, and cylinder, with 6 possible colors to choose from. Object sizes are sampled uniformly between $[0.56, 0.66]$ units in the MuJoCo \citep{mujoco} physics world. All objects are placed on the $z=0$ plane, with a range of $[-2, 2]$ along both $x$-axis and $y$-axis. We randomly sample 30 cameras for each scene. They are placed at a distance of $3$ from the origin, but do not necessarily point to the origin. The camera pitch is sampled between $[\pi/7, \pi/6]$ so that the camera is always above the $z=0$ plane. The camera yaw is sampled between $[-\pi, \pi]$.

\textbf{Multi-Shepard-Metzler.}
We generate the Shepard-Metzler objects as described in GQN \citep{gqn}. Each object consists of 5 cubes with edge length $0.8$. Each cube is randomly colored, with hue between $[0, 1]$, saturation between $[0.75, 1]$, and value equal to $1$. Like the Shapes data set, all objects are placed on the $z=0$ plane, with a range of $[-3, 3]$ along both $x$-axis and $y$-axis. We randomly sample 30 cameras for each scene and place them at a distance of $12$ from the origin. They all point to the origin. The camera pitch is sampled between $[-\frac{5}{12}\pi, \frac{5}{12}\pi]$, and the yaw is sampled between $[-\pi, \pi]$.

\section{ROOTS Implementation Details}\label{module_details}
In this section, we introduce the key building blocks for implementing ROOTS.

\textbf{Context Encoder.} The context encoder is modified based on the `tower' representation architecture in GQN \citep{gqn}. It encodes each pair of context image and the corresponding viewpoint into a vector. Summation is applied over the context encodings to obtain the order-invariant representation $\bpsi$.



\textbf{Object-Level Context Encoder.} The object-level context encoder is also an adaptation of the `tower' representation architecture, but takes the extracted object-level context $\gC_n$ as input.


\textbf{{ConvDRAW.}} We use ConvDRAW \citep{gregor2016towards} to infer the prior and posterior distributions of the latent variables. To render the objects and the background, we use a deterministic version of ConvDRAW (i.e., without sampling). In the following, we describe one rollout step (denoted $l$) of ConvDRAW used in ROOTS generative and inference processes, respectively. We provide detailed configurations of each ConvDRAW module in Table~\ref{spec_convdraw}.

\begin{itemize}
    \item Generative Process:
    \begin{align*}
    (\mathbf{h}^{(l+1)}_p, \mathbf{c}^{(l+1)}_p) &\leftarrow \text{ConvLSTM}_{\theta}{(\bpsi_\gC, \rvz^{(l)}, \mathbf{h}^{(l)}_p, \mathbf{c}^{(l)}_p}) \\
    \mathbf{z}^{(l+1)} &\sim \text{StatisNet}_{\theta}{(\mathbf{h}^{(l+1)}_p)}
    \end{align*}
    \item Inference Process:
    \begin{align*}
    (\mathbf{h}^{(l+1)}_p, \mathbf{c}^{(l+1)}_p) &\leftarrow \text{ConvLSTM}_{\theta}{(\bpsi_\gC, \rvz^{(l)},  \mathbf{h}^{(l)}_p, \mathbf{c}^{(l)}_p}) \\
    (\mathbf{h}^{(l+1)}_q, \mathbf{c}^{(l+1)}_q) &\leftarrow \text{ConvLSTM}_{\phi}{(\bpsi_\gC, \bpsi_\gQ, \mathbf{h}^{(l)}_p, \mathbf{h}^{(l)}_q, \mathbf{c}^{(l)}_q}) \\
    \mathbf{z}^{(l+1)} &\sim \text{StatisNet}_{\theta}{(\mathbf{h}^{(l+1)}_q)}
\end{align*}
\end{itemize}
Here, $\bpsi_\gC$ and $\bpsi_\gQ$ are order-invariant encodings of contexts and queries respectively, and $\mathbf{z}^{(l+1)}$ is the sampled latent at the $(l+1)$-th step. The prior module is denoted by subscript $p$, with learnable parameters $\theta$, and the posterior module is denoted by subscript $q$, with learnable parameters $\phi$. StatisNet maps hidden states to distribution parameters, and will be explained in the following. ConvLSTM replaces the fully-connected layers in LSTM \citep{hochreiter1997long} by convolutional layers.


\begin{table}[t]
    \centering
    \begin{tabular}{cccc}
	    \toprule
	    Module Name & Rollout Steps & Hidden Size\\ 
	    \midrule
	    $\rvz^{\bg}$      & 2 & 128    \\
	    $\rvz^{\where}$  & 2 & 128    \\
	    $\rvz^{\what}$   & 4 & 128    \\
	    Object Renderer  & 4 & 128    \\
	    Background Renderer  & 2 & 128    \\
	    \bottomrule
    \end{tabular}
    \caption{Configuration of ConvDRAW modules.}
    \label{spec_convdraw}
\end{table}

\textbf{Sufficient Statistic Network.} The Sufficient Statistic Network (StatisNet) outputs sufficient statistics for pre-defined distributions, e.g., $\mu$ and $\sigma$ for Gaussian distributions, and $\rho$ for Bernoulli distributions. We list the configuration of all Sufficient Statistic Networks in Table \ref{stat}. For $\rvz_{n}^{\where}$, $\rvz_{n}^{\what}$, and $\rvz^{\bg}$, we use ConvDraw to learn the sufficient statistics. For $\rz_n^{\pres}$, two ConvBlocks are first used to extract features, and then a third ConvBlock combines the features and outputs the parameter of the Bernoulli distribution. GN denotes group normalization \citep{wu2018group}, and CELU denotes continuously differentiable exponential linear units \citep{barron2017continuously}.  

\begin{table}[t]
\vskip 0.2in
	\begin{center}
		\begin{adjustbox}{max width=\textwidth}
		\begin{tabular}{ccc}
			\toprule
			\multicolumn{3}{c}{Object-Level Latent Variables}\\
			\midrule
			$\rvz^{\where}$ & $\rvz^{\what}$  &\\ 
     		\cmidrule(lr){1-1} \cmidrule(lr){2-2}
     		
			\multicolumn{1}{l}{Conv3D(128, 3, 1, GN, CELU)} & \multicolumn{1}{l}{Conv3D(128, 1, 1, GN, CELU)} &  \\ 
			
			\multicolumn{1}{l}{Conv3D(64, 1, 1, GN, CELU)} & \multicolumn{1}{l}{Conv3D(64, 1, 1, GN, CELU)} & \\ 
			
			\multicolumn{1}{l}{Conv3D(32, 1, 1, GN, CELU)} & \multicolumn{1}{l}{Conv3D(32, 1, 1, GN, CELU)} & \\ 
			
			\multicolumn{1}{l}{Conv3D(3 $\times$ 2, 1, 1, GN, CELU)} & \multicolumn{1}{l}{Conv3D(4 $\times$ 2, 1, 1, GN, CELU)} & \\ 
			\midrule
			\multicolumn{3}{c}{$\rz^{\pres}$} \\
			\cmidrule(lr){1-3}
			
			ConvBlock 1 & ConvBlock 2 & ConvBlock 3\\
			\cmidrule(lr){1-1} \cmidrule(lr){2-2} \cmidrule(lr){3-3}
			
			\multicolumn{1}{l}{Conv3D(256, 3, 1, GN, CELU)} & \multicolumn{1}{l}{Conv3D(256, 1, 1, GN, CELU)} &
			\multicolumn{1}{l}{Conv3D(128, 3, 1, GN, CELU)} \\
			& \multicolumn{1}{l}{Conv3D(256, 1, 1, GN, CELU)} &
			\multicolumn{1}{l}{Conv3D(64, 1, 1, GN, CELU)} \\
			& & \multicolumn{1}{l}{Conv3D(1, 1, 1, GN, CELU)}\\
			\midrule
			\midrule
			\multicolumn{3}{c}{Scene-Level Latent Variables}\\
			\midrule
			$\rvz^{\bg}$ & & \\
			
			\cmidrule(lr){1-1}
			
			Conv3D(1 $\times$ 2, 1, 1, GN, CELU)  & &\\
			
			\bottomrule
		\end{tabular}
		\end{adjustbox}
	\end{center}
	\caption{Configuration of Sufficient Statistic Networks.}
	\label{stat}
\end{table}

\section{Downstream Task Details}
We use the 3-5 Shapes data set for the downstream tasks. To generate the ground-truth labels, we assign a label to each object based on its type and color. There are $3$ different types and $6$ different colors in total, thus the label value for one object lies in the range from $0$ to $17$. We split the data set into training set, validation set, and test set of size $50$K, $5$K, and $5$K, respectively. During training, $10$ to $20$ randomly sampled context observations are provided for both GQN and ROOTS to learn representations of a 3D scene. All latent variables are sampled from the learned priors.

\textbf{Retrieve Object.} To predict the correct class of the object that lies closest to a given point in the 3D space, the classifier first encodes the scene representation $\hat{\bm{r}}$ into a vector, and then uses MLP to predict class probabilities. For GQN, we concatenate the scene embedding and scene latent representation together, that is, $\hat{\bm{r}} = [\bm{r}_\gC, \rvz]$. For ROOTS, we use the object-level representations, that is,  $\hat{\bm{r}} = \{\bm{r}_n^\att, \rvz_n^\mathrm{3D}\}_{n=1}^N$, where $\bm{r}_n^\att$ is an order-invariant encoding of object-level context ${\gC}_n$. For 2D version of this task, we provide $\rvv_q$ as an additional input to the classifier. The structure of the scene representation encoder is specified in Table \ref{spec_downstream}. We use three linear layers for the classifier network, as listed in Table \ref{spec_downstream}. For both versions of this task, the number of classes $N_{\text{cls}} = 18$.

\textbf{Find Pair.} The object-level representation provided by ROOTS naturally allows us to extract pair-wise relationships by using a graph net \citep{battaglia2018relational,velivckovic2017graph}. Specifically, we use the object-level representation as the node feature. We then extract edge features for each pair of objects using shared MLPs (Edge Encoder). The edge features are pooled into a single vector using attention, and fed into another MLP (Graph Encoder) to produce the final classification result. For GQN, we encode the scene representation provided by GQN into a GVFM, so that object-specific features can be split into individual cells. We then treat each cell as a node feature, and apply a similar graph net for classification. The Edge Encoder and Graph Encoder of the graph net is specified in Table \ref{spec_downstream}. We use the same classifier structure as the Retrieve Object task. For this task, the number of classes $N_{\text{cls}} = 35$.

\begin{table}[t]
	\begin{center}
	\begin{adjustbox}{max width=\textwidth}
		\begin{tabular}{ccc}
			\toprule
			\multicolumn{2}{c}{Scene Representation Encoder} & \multirow{2}{*}{GQN-GVFM} \\
			\cmidrule(lr){1-2}
			GQN & ROOTS & \\ 
			\cmidrule(lr){1-1} \cmidrule(lr){2-2} \cmidrule(lr){3-3}
			
			Conv(256, 2, 2, GN, CELU)  & Conv(256, 1, 1, GN, CELU) &  Conv(256, 5, 3) \\ 
			Conv(256, 2, 2, GN, CELU)  & Conv(256, 3, 1, GN, CELU) &  Conv(128, 3, 2) \\	
			Conv(256, 4, 4, GN, CELU)  & Conv(256, 3, 1, GN, CELU) &  Conv(256, 3, 1) \\	
			 & & ConvTrans3D(256, [3,1,3], 1) \\
			
			\midrule
			\midrule
			\multicolumn{2}{c}{Graph Network} & \multirow{2}{*}{Classifier Network} \\
			\cmidrule(lr){1-2} 
			Edge Encoder & Graph Encoder \\
			\cmidrule(lr){1-1} \cmidrule(lr){2-2} \cmidrule(lr){3-3}
			
			MLP(256, 256, 256+1)  & MLP(256, 256) &  MLP(128, 64, $N_{\text{cls}}$) \\
			\bottomrule
		\end{tabular}
	\end{adjustbox}
	\end{center}
	\caption{Configuration of downstream task networks.}
	\label{spec_downstream}
\end{table}

\section{CGQN Baseline}

We use the `tower' representation architecture in GQN \citep{gqn} to encode the context $\gC$ into a scene-level representation $\bm{r}_\gC$ of size $16 \times 16 \times 256$. We then follow the CGQN paper \citep{cgqn} and use ConvDraw \citep{gregor2016towards} to sample $\rvz$ from $\bm{r}_\gC$ and decode the target image $\rvx_q$ from $\rvz$ and query viewpoint $\rvv_q$. The main hyperparameters are ConvDraw steps (denoted $L$) and the number of $\rvz$ channels (denoted $c$) that are sampled at each ConvDraw step. Generally, larger values of $L$ and $c$ give stronger model capacity, but take more computational resource. We choose $L=12$ and $c=4$ for the Shapes data set and the ShapeNet arrangement data set, and $L=16$ and $c=8$ for the Multi-Shepard-Metzler data set.

We trained the CGQN baseline using Adam \citep{adam} with learning rates chosen from $\{1 \times 10^{-3}, 3 \times 10^{-4}, 1 \times 10^{-4}, 3 \times 10^{-5}\}$. We found the learning rate of $3 \times 10^{-4}$ worked best for the Shapes data set and the Multi-Shepard-Metzler data set, while for the ShapeNet arrangement data set, the learning rate of $1 \times 10^{-4}$ worked best. For fair comparison of NLL, we used a fixed pixel-variance $\sigma^2$ of 0.09 during training for both ROOTS and CGQN. To be consistent with the pixel-variance annealing strategy used in CGQN, we multiplied the KL divergence by a value $\beta$ that is linearly annealed from $20$ to $5$ at the start of training.

\section{IODINE Baseline}\label{app:iodine}
\begin{figure}[hb]
    \centering
    \includegraphics[width=0.56\textwidth]{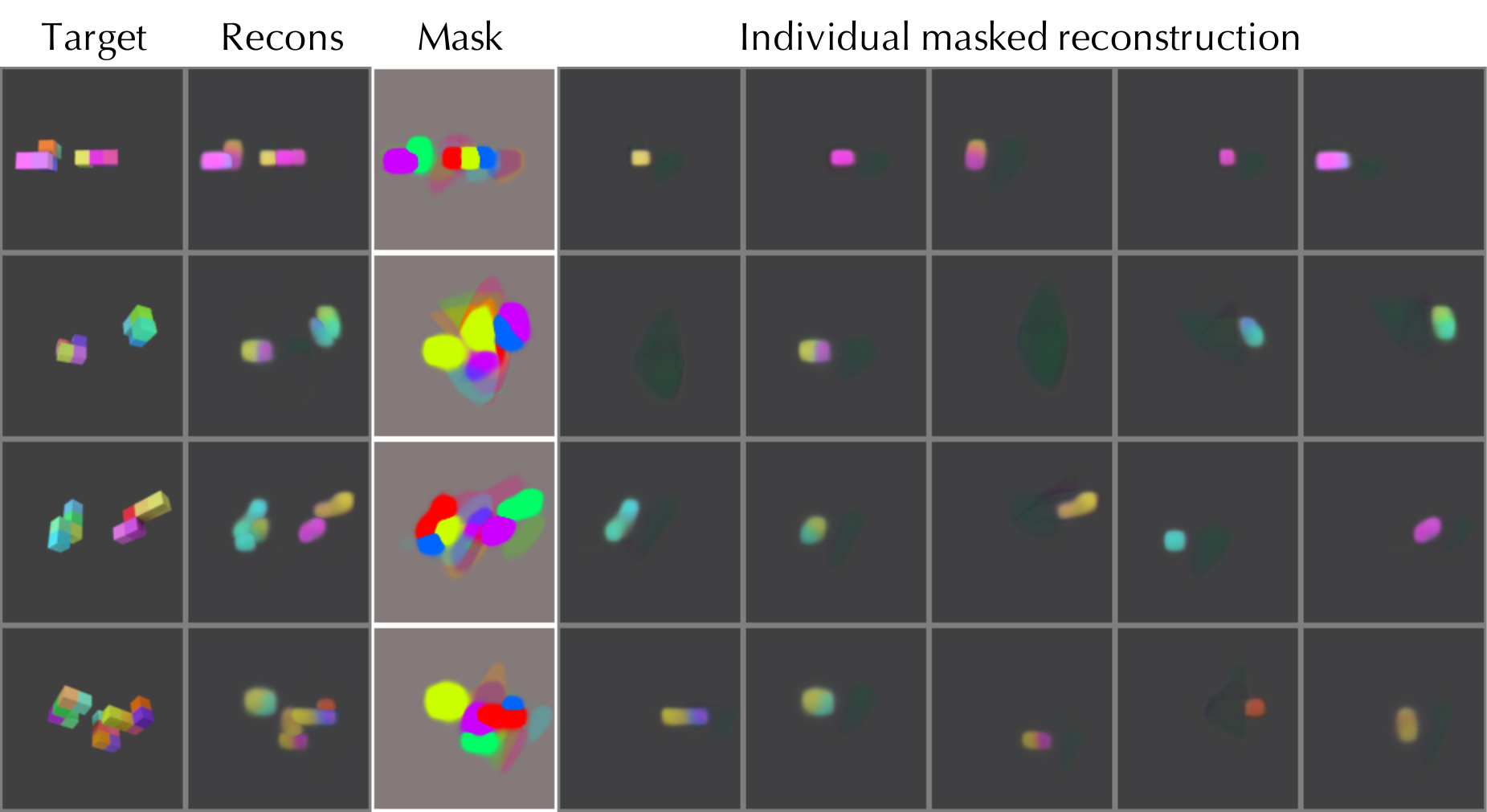}
    \caption{Scene decomposition results of IODINE with Spatial Broadcast decoder on the Multi-Shepard-Metzler data set.}
    \label{iodine_spatial}
\end{figure}

\begin{figure}[ht]
    \centering
    \includegraphics[width=0.56\textwidth]{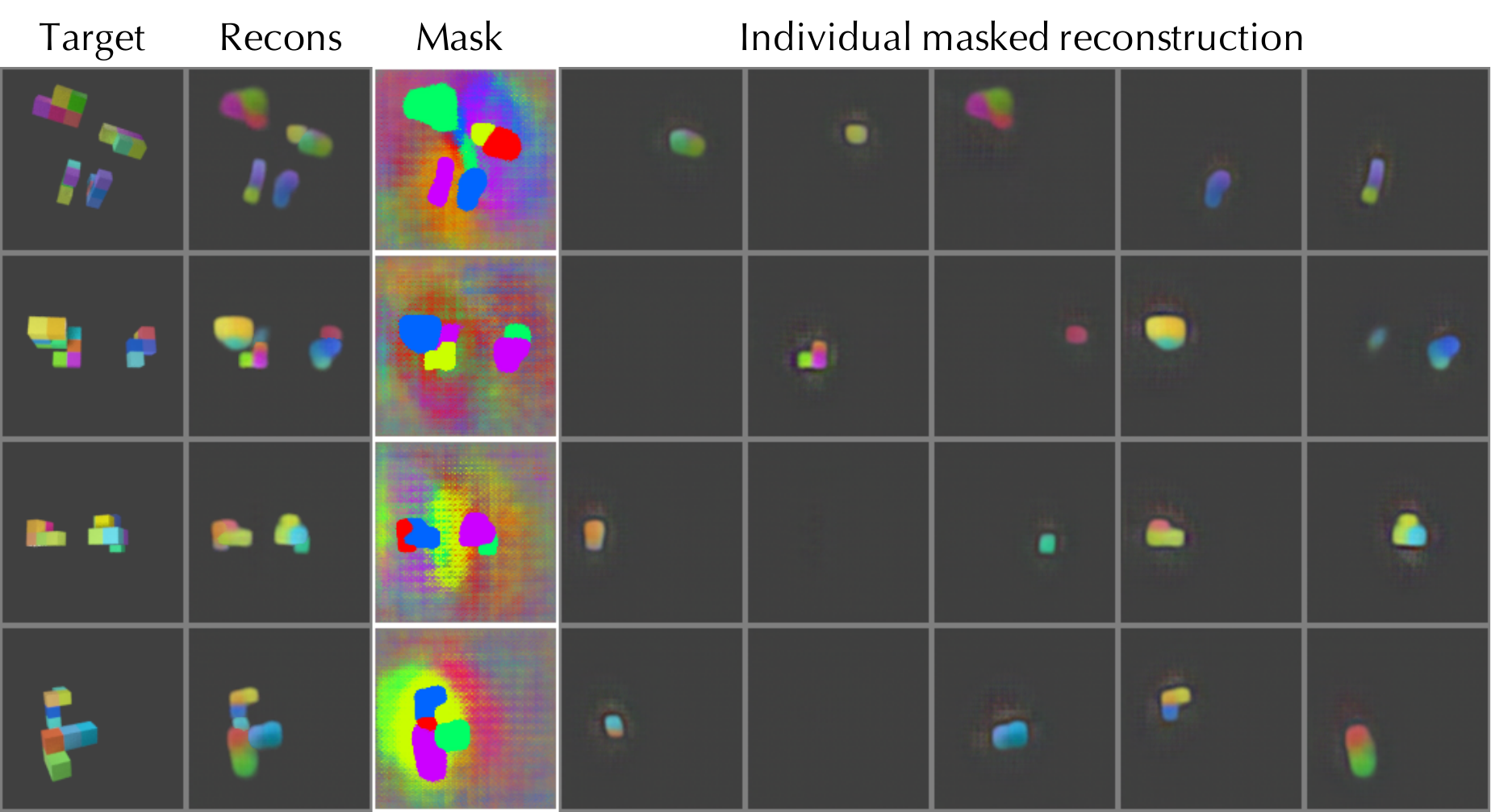}
    \caption{Scene decomposition results of IODINE with deconvolution-based decoder on the Multi-Shepard-Metzler data set.}
    \label{iodine_deconv}
\end{figure}

We use the implementation from DeepMind for IODINE \citep{iodine}. We adjusted the output standard deviation $\sigma$ and the slot number $K$ for best performance. We tried values for $\sigma$ in the range of [0.1, 0.3], and found that the model tends to be unstable with smaller $\sigma$ values.
We varied $K$ from $4$ to $8$. The final values used for the Shapes data set are listed in Table \ref{iodine}. For the Multi-Shepard-Metzler data set, we found that IODINE tends to segment objects based on color similarity, thereby splitting a single object into multiple slots. We show two examples in \Figref{iodine_spatial} and \Figref{iodine_deconv}, where the Spatial Broadcast decoder \citep{watters2019spatial} and deconvolution-based decoder are used respectively. We use $K=5$ and $\sigma=0.25$ in both settings.

\begin{table}[t]
    \centering
    \begin{tabular}{cccc}
    \toprule
	
	Data Set & 1-3 Shapes & 2-4 Shapes & 3-5 Shapes   \\ 
	
	\midrule
	$\sigma$      & 0.3  & 0.3  & 0.25   \\
	$K$           & 4    & 7    & 8   \\
	
	\bottomrule
    \end{tabular}
    \caption{Hyperparameters for IODINE.}
    \label{iodine}
\end{table}

\newpage
\bibliography{refs_ahn_,refs}

\begin{thebibliography}{75}
\providecommand{\natexlab}[1]{#1}
\providecommand{\url}[1]{\texttt{#1}}
\expandafter\ifx\csname urlstyle\endcsname\relax
  \providecommand{\doi}[1]{doi: #1}\else
  \providecommand{\doi}{doi: \begingroup \urlstyle{rm}\Url}\fi

\bibitem[Achlioptas et~al.(2018)Achlioptas, Diamanti, Mitliagkas, and
  Guibas]{achlioptas2018learning}
Panos Achlioptas, Olga Diamanti, Ioannis Mitliagkas, and Leonidas Guibas.
\newblock Learning representations and generative models for 3{D} point clouds.
\newblock In \emph{International Conference on Machine Learning}, 2018.

\bibitem[Babaeizadeh et~al.(2018)Babaeizadeh, Finn, Erhan, Campbell, and
  Levine]{babaeizadeh2018stochastic}
Mohammad Babaeizadeh, Chelsea Finn, Dumitru Erhan, Roy~H. Campbell, and Sergey
  Levine.
\newblock Stochastic variational video prediction.
\newblock In \emph{International Conference on Learning Representations}, 2018.

\bibitem[Bahdanau et~al.(2019)Bahdanau, Murty, Noukhovitch, Nguyen, de~Vries,
  and Courville]{bahdanau2018systematic}
Dzmitry Bahdanau, Shikhar Murty, Michael Noukhovitch, Thien~Huu Nguyen, Harm
  de~Vries, and Aaron Courville.
\newblock Systematic generalization: What is required and can it be learned?
\newblock In \emph{International Conference on Learning Representations}, 2019.

\bibitem[Barron(2017)]{barron2017continuously}
Jonathan~T. Barron.
\newblock Continuously differentiable exponential linear units.
\newblock \emph{arXiv preprint arXiv:1704.07483}, 2017.

\bibitem[Battaglia et~al.(2018)Battaglia, Hamrick, Bapst, Sanchez-Gonzalez,
  Zambaldi, Malinowski, Tacchetti, Raposo, Santoro, Faulkner, Gulcehre, Song,
  Ballard, Gilmer, Dahl, Vaswani, Allen, Nash, Langston, Dyer, Heess, Wierstra,
  Kohli, Botvinick, Vinyals, Li, and Pascanu]{battaglia2018relational}
Peter~W. Battaglia, Jessica~B. Hamrick, Victor Bapst, Alvaro Sanchez-Gonzalez,
  Vinicius Zambaldi, Mateusz Malinowski, Andrea Tacchetti, David Raposo, Adam
  Santoro, Ryan Faulkner, Caglar Gulcehre, Francis Song, Andrew Ballard, Justin
  Gilmer, George Dahl, Ashish Vaswani, Kelsey Allen, Charles Nash, Victoria
  Langston, Chris Dyer, Nicolas Heess, Daan Wierstra, Pushmeet Kohli, Matt
  Botvinick, Oriol Vinyals, Yujia Li, and Razvan Pascanu.
\newblock Relational inductive biases, deep learning, and graph networks.
\newblock \emph{arXiv preprint arXiv:1806.01261}, 2018.

\bibitem[{Blender Online Community}(2017)]{blender}
{Blender Online Community}.
\newblock \emph{Blender---a 3{D} modelling and rendering package}.
\newblock Blender Foundation, Blender Institute, Amsterdam, 2017.

\bibitem[Bottou(2014)]{bottou2014machine}
L{\'e}on Bottou.
\newblock From machine learning to machine reasoning.
\newblock \emph{Machine Learning}, 94\penalty0 (2):\penalty0 133--149, 2014.

\bibitem[Burgess et~al.(2019)Burgess, Matthey, Watters, Kabra, Higgins,
  Botvinick, and Lerchner]{monet}
Christopher~P. Burgess, Loic Matthey, Nicholas Watters, Rishabh Kabra, Irina
  Higgins, Matt Botvinick, and Alexander Lerchner.
\newblock {MONet}: Unsupervised scene decomposition and representation.
\newblock \emph{arXiv preprint arXiv:1901.11390}, 2019.

\bibitem[Chang et~al.(2015)Chang, Funkhouser, Guibas, Hanrahan, Huang, Li,
  Savarese, Savva, Song, Su, Xiao, Yi, and Yu]{shapenet2015}
Angel~X. Chang, Thomas Funkhouser, Leonidas Guibas, Pat Hanrahan, Qixing Huang,
  Zimo Li, Silvio Savarese, Manolis Savva, Shuran Song, Hao Su, Jianxiong Xiao,
  Li~Yi, and Fisher Yu.
\newblock {ShapeNet}: {An} information-rich 3{D} model repository.
\newblock \emph{arXiv preprint arXiv:1512.03012}, 2015.

\bibitem[Cheng et~al.(2018)Cheng, Wang, and Fragkiadaki]{cheng2018geometry}
Ricson Cheng, Ziyan Wang, and Katerina Fragkiadaki.
\newblock Geometry-aware recurrent neural networks for active visual
  recognition.
\newblock In \emph{Advances in Neural Information Processing Systems}, pages
  5081--5091, 2018.

\bibitem[Choy et~al.(2016)Choy, Xu, Gwak, Chen, and Savarese]{choy20163d}
Christopher~B. Choy, Danfei Xu, JunYoung Gwak, Kevin Chen, and Silvio Savarese.
\newblock {3D-R2N2}: {A} unified approach for single and multi-view 3{D} object
  reconstruction.
\newblock In \emph{Proceedings of the European Conference on Computer Vision},
  2016.

\bibitem[Crawford and Pineau(2019)]{spair}
Eric Crawford and Joelle Pineau.
\newblock Spatially invariant unsupervised object detection with convolutional
  neural networks.
\newblock In \emph{Thirty-Third AAAI Conference on Artificial Intelligence},
  2019.

\bibitem[Crawford and Pineau(2020)]{srn_silot}
Eric Crawford and Joelle Pineau.
\newblock Learning 3{D} object-oriented world models from unlabeled videos.
\newblock \emph{Workshop on Object-Oriented Learning at ICML}, 2020.

\bibitem[Du et~al.(2018)Du, Liu, Basevi, Leonardis, Freeman, Tenenbaum, and
  Wu]{du2018learning}
Yilun Du, Zhijian Liu, Hector Basevi, Ales Leonardis, Bill Freeman, Josh
  Tenenbaum, and Jiajun Wu.
\newblock Learning to exploit stability for 3{D} scene parsing.
\newblock In \emph{Advances in Neural Information Processing Systems}, pages
  1726--1736, 2018.

\bibitem[Dupont et~al.(2020)Dupont, Bautista, Colburn, Sankar, Guestrin,
  Susskind, and Shan]{enr}
Emilien Dupont, Miguel~Angel Bautista, Alex Colburn, Aditya Sankar, Carlos
  Guestrin, Josh Susskind, and Qi~Shan.
\newblock Equivariant neural rendering.
\newblock In \emph{International Conference on Machine Learning}, 2020.

\bibitem[Ehrhardt et~al.(2020)Ehrhardt, Groth, Monszpart, Engelcke, Posner,
  Mitra, and Vedaldi]{ehrhardt2020relate}
Sebastien Ehrhardt, Oliver Groth, Aron Monszpart, Martin Engelcke, Ingmar
  Posner, Niloy Mitra, and Andrea Vedaldi.
\newblock {RELATE}: {P}hysically plausible multi-object scene synthesis using
  structured latent spaces.
\newblock \emph{arXiv preprint arXiv:2007.01272}, 2020.

\bibitem[Engelcke et~al.(2020)Engelcke, Kosiorek, Parker~Jones, and
  Posner]{genesis}
Martin Engelcke, Adam~R. Kosiorek, Oiwi Parker~Jones, and Ingmar Posner.
\newblock {GENESIS}: {G}enerative scene inference and sampling of
  object-centric latent representations.
\newblock In \emph{International Conference on Learning Representations}, 2020.

\bibitem[Eslami et~al.(2016)Eslami, Heess, Weber, Tassa, Szepesvari, and
  Hinton]{air}
S.M.~Ali Eslami, Nicolas Heess, Theophane Weber, Yuval Tassa, David Szepesvari,
  and Geoffrey~E. Hinton.
\newblock {Attend, Infer, Repeat}: {F}ast scene understanding with generative
  models.
\newblock In \emph{Advances in Neural Information Processing Systems}, pages
  3225--3233, 2016.

\bibitem[Eslami et~al.(2018)Eslami, Rezende, Besse, Viola, Morcos, Garnelo,
  Ruderman, Rusu, Danihelka, Gregor, Reichert, Buesing, Weber, Vinyals,
  Rosenbaum, Rabinowitz, King, Hillier, Botvinick, Wierstra, Kavukcuoglu, and
  Hassabis]{gqn}
S.M.~Ali Eslami, Danilo~Jimenez Rezende, Frederic Besse, Fabio Viola, Ari~S.
  Morcos, Marta Garnelo, Avraham Ruderman, Andrei~A. Rusu, Ivo Danihelka, Karol
  Gregor, David~P. Reichert, Lars Buesing, Theophane Weber, Oriol Vinyals, Dan
  Rosenbaum, Neil Rabinowitz, Helen King, Chloe Hillier, Matt Botvinick, Daan
  Wierstra, Koray Kavukcuoglu, and Demis Hassabis.
\newblock Neural scene representation and rendering.
\newblock \emph{Science}, 360\penalty0 (6394):\penalty0 1204--1210, 2018.

\bibitem[Everingham et~al.(2010)Everingham, Van~Gool, Williams, Winn, and
  Zisserman]{everingham2010pascal}
Mark Everingham, Luc Van~Gool, Christopher~K.I. Williams, John Winn, and Andrew
  Zisserman.
\newblock The {P\textsc{ascal} Visual Object Classes} ({VOC}) challenge.
\newblock \emph{International Journal of Computer Vision}, 88\penalty0
  (2):\penalty0 303--338, 2010.

\bibitem[Goodfellow et~al.(2014)Goodfellow, Pouget-Abadie, Mirza, Xu,
  Warde-Farley, Ozair, Courville, and Bengio]{gan}
Ian~J. Goodfellow, Jean Pouget-Abadie, Mehdi Mirza, Bing Xu, David
  Warde-Farley, Sherjil Ozair, Aaron Courville, and Yoshua Bengio.
\newblock Generative adversarial nets.
\newblock In \emph{Advances in Neural Information Processing Systems}, page
  2672–2680, 2014.

\bibitem[Greff et~al.(2017)Greff, van Steenkiste, and Schmidhuber]{nem}
Klaus Greff, Sjoerd van Steenkiste, and J{\"u}rgen Schmidhuber.
\newblock Neural expectation maximization.
\newblock In \emph{Advances in Neural Information Processing Systems}, pages
  6691--6701, 2017.

\bibitem[Greff et~al.(2019)Greff, Kaufmann, Kabra, Watters, Burgess, Zoran,
  Matthey, Botvinick, and Lerchner]{iodine}
Klaus Greff, Rapha{\"e}l~Lopez Kaufmann, Rishab Kabra, Nick Watters, Chris
  Burgess, Daniel Zoran, Loic Matthey, Matthew Botvinick, and Alexander
  Lerchner.
\newblock Multi-object representation learning with iterative variational
  inference.
\newblock In \emph{International Conference on Machine Learning}, 2019.

\bibitem[Gregor et~al.(2016)Gregor, Besse, Rezende, Danihelka, and
  Wierstra]{gregor2016towards}
Karol Gregor, Frederic Besse, Danilo~Jimenez Rezende, Ivo Danihelka, and Daan
  Wierstra.
\newblock Towards conceptual compression.
\newblock In \emph{Advances In Neural Information Processing Systems}, pages
  3549--3557, 2016.

\bibitem[Hartley and Zisserman(2003)]{hartley2003multiple}
Richard Hartley and Andrew Zisserman.
\newblock \emph{Multiple View Geometry in Computer Vision}.
\newblock Cambridge University Press, 2003.

\bibitem[Higgins et~al.(2017)Higgins, Matthey, Pal, Burgess, Glorot, Botvinick,
  Mohamed, and Lerchner]{betavae}
Irina Higgins, Lo{\"\i}c Matthey, Arka Pal, Christopher Burgess, Xavier Glorot,
  Matthew~M. Botvinick, Shakir Mohamed, and Alexander Lerchner.
\newblock $\beta$-{VAE}: {L}earning basic visual concepts with a constrained
  variational framework.
\newblock In \emph{International Conference on Learning Representations}, 2017.

\bibitem[Hochreiter and Schmidhuber(1997)]{hochreiter1997long}
Sepp Hochreiter and J{\"u}rgen Schmidhuber.
\newblock Long short-term memory.
\newblock \emph{Neural computation}, 9\penalty0 (8):\penalty0 1735--1780, 1997.

\bibitem[Hood and Santos(2009)]{hood2009origins}
Bruce Hood and Laurie Santos.
\newblock \emph{The Origins of Object Knowledge}.
\newblock Oxford University Press, 2009.

\bibitem[H{\o}ydal et~al.(2019)H{\o}ydal, Skyt{\o}en, Andersson, Moser, and
  Moser]{hoydal2019object}
{\O}yvind~Arne H{\o}ydal, Emilie~Ranheim Skyt{\o}en, Sebastian~Ola Andersson,
  May-Britt Moser, and Edvard~I. Moser.
\newblock Object-vector coding in the medial entorhinal cortex.
\newblock \emph{Nature}, 568\penalty0 (7752):\penalty0 400--404, 2019.

\bibitem[Huang et~al.(2018)Huang, Qi, Xiao, Zhu, Wu, and
  Zhu]{huang2018cooperative}
Siyuan Huang, Siyuan Qi, Yinxue Xiao, Yixin Zhu, Ying~Nian Wu, and Song-Chun
  Zhu.
\newblock Cooperative holistic scene understanding: {U}nifying 3{D} object,
  layout, and camera pose estimation.
\newblock In \emph{Advances in Neural Information Processing Systems}, pages
  207--218, 2018.

\bibitem[Hubert and Arabie(1985)]{hubert1985comparing}
Lawrence Hubert and Phipps Arabie.
\newblock Comparing partitions.
\newblock \emph{Journal of Classification}, 2\penalty0 (1):\penalty0 193--218,
  1985.

\bibitem[Jaderberg et~al.(2015)Jaderberg, Simonyan, Zisserman, and
  Kavukcuoglu]{spatial_transformer}
Max Jaderberg, Karen Simonyan, Andrew Zisserman, and Koray Kavukcuoglu.
\newblock Spatial transformer networks.
\newblock In \emph{Advances in Neural Information Processing Systems}, pages
  2017--2025, 2015.

\bibitem[Jang et~al.(2017)Jang, Gu, and Poole]{jang2016categorical}
Eric Jang, Shixiang Gu, and Ben Poole.
\newblock Categorical reparametrization with {G}umbel-{S}oftmax.
\newblock In \emph{International Conference on Learning Representations}, 2017.

\bibitem[Kahneman et~al.(1992)Kahneman, Treisman, and Gibbs]{objectfiles}
Daniel Kahneman, Anne Treisman, and Brian~J. Gibbs.
\newblock The reviewing of object files: {O}bject-specific integration of
  information.
\newblock \emph{Cognitive Psychology}, 24\penalty0 (2):\penalty0 175--219,
  1992.

\bibitem[Kanazawa et~al.(2018)Kanazawa, Tulsiani, Efros, and
  Malik]{kanazawa2018learning}
Angjoo Kanazawa, Shubham Tulsiani, Alexei~A. Efros, and Jitendra Malik.
\newblock Learning category-specific mesh reconstruction from image
  collections.
\newblock In \emph{Proceedings of the European Conference on Computer Vision},
  2018.

\bibitem[Kar et~al.(2017)Kar, H{\"a}ne, and Malik]{kar2017learning}
Abhishek Kar, Christian H{\"a}ne, and Jitendra Malik.
\newblock Learning a multi-view stereo machine.
\newblock In \emph{Advances in Neural Information Processing Systems}, pages
  365--376, 2017.

\bibitem[Kato et~al.(2018)Kato, Ushiku, and Harada]{kato2018neural}
Hiroharu Kato, Yoshitaka Ushiku, and Tatsuya Harada.
\newblock Neural 3{D} mesh renderer.
\newblock In \emph{Proceedings of the IEEE/CVF Conference on Computer Vision
  and Pattern Recognition}, 2018.

\bibitem[Kingma and Ba(2015)]{adam}
Diederik~P. Kingma and Jimmy Ba.
\newblock Adam: {A} method for stochastic optimization.
\newblock In \emph{International Conference on Learning Representations}, 2015.

\bibitem[Kingma and Welling(2014)]{vae}
Diederik~P. Kingma and Max Welling.
\newblock Auto-encoding variational {B}ayes.
\newblock In \emph{International Conference on Learning Representations}, 2014.

\bibitem[Kumar et~al.(2018)Kumar, Eslami, Rezende, Garnelo, Viola, Lockhart,
  and Shanahan]{cgqn}
Ananya Kumar, S.M.~Ali Eslami, Danilo~Jimenez Rezende, Marta Garnelo, Fabio
  Viola, Edward Lockhart, and Murray Shanahan.
\newblock Consistent generative query networks.
\newblock \emph{arXiv preprint arXiv:1807.02033}, 2018.

\bibitem[Lake et~al.(2017)Lake, Ullman, Tenenbaum, and
  Gershman]{lake2017building}
Brenden~M. Lake, Tomer~D. Ullman, Joshua~B. Tenenbaum, and Samuel~J. Gershman.
\newblock Building machines that learn and think like people.
\newblock \emph{Behavioral and Brain Sciences}, 40, 2017.

\bibitem[Liao et~al.(2020)Liao, Schwarz, Mescheder, and
  Geiger]{liao2019towards}
Yiyi Liao, Katja Schwarz, Lars Mescheder, and Andreas Geiger.
\newblock Towards unsupervised learning of generative models for 3{D}
  controllable image synthesis.
\newblock In \emph{Proceedings of the IEEE/CVF Conference on Computer Vision
  and Pattern Recognition}, 2020.

\bibitem[Lin et~al.(2020)Lin, Wu, Peri, Sun, Singh, Deng, Jiang, and
  Ahn]{space}
Zhixuan Lin, Yi-Fu Wu, Skand~Vishwanath Peri, Weihao Sun, Gautam Singh, Fei
  Deng, Jindong Jiang, and Sungjin Ahn.
\newblock {SPACE}: Unsupervised object-oriented scene representation via
  spatial attention and decomposition.
\newblock In \emph{International Conference on Learning Representations}, 2020.

\bibitem[Locatello et~al.(2020)Locatello, Weissenborn, Unterthiner, Mahendran,
  Heigold, Uszkoreit, Dosovitskiy, and Kipf]{slotattention}
Francesco Locatello, Dirk Weissenborn, Thomas Unterthiner, Aravindh Mahendran,
  Georg Heigold, Jakob Uszkoreit, Alexey Dosovitskiy, and Thomas Kipf.
\newblock Object-centric learning with slot attention.
\newblock \emph{arXiv preprint arXiv:2006.15055}, 2020.

\bibitem[Maddison et~al.(2017)Maddison, Mnih, and Teh]{maddison2016concrete}
Chris~J. Maddison, Andriy Mnih, and Yee~Whye Teh.
\newblock The {Concrete} distribution: {A} continuous relaxation of discrete
  random variables.
\newblock In \emph{International Conference on Learning Representations}, 2017.

\bibitem[Martin(2007)]{martin2007representation}
Alex Martin.
\newblock The representation of object concepts in the brain.
\newblock \emph{Annual Review of Psychology}, 58:\penalty0 25--45, 2007.

\bibitem[Maturana and Scherer(2015)]{maturana2015voxnet}
Daniel Maturana and Sebastian Scherer.
\newblock Vox{N}et: {A} 3{D} convolutional neural network for real-time object
  recognition.
\newblock In \emph{2015 IEEE/RSJ International Conference on Intelligent Robots
  and Systems}, pages 922--928. IEEE, 2015.

\bibitem[Mildenhall et~al.(2020)Mildenhall, Srinivasan, Tancik, Barron,
  Ramamoorthi, and Ng]{nerf}
Ben Mildenhall, Pratul~P. Srinivasan, Matthew Tancik, Jonathan~T. Barron, Ravi
  Ramamoorthi, and Ren Ng.
\newblock Ne{RF}: Representing scenes as neural radiance fields for view
  synthesis.
\newblock In \emph{Proceedings of the European Conference on Computer Vision},
  2020.

\bibitem[Nguyen-Phuoc et~al.(2019)Nguyen-Phuoc, Li, Theis, Richardt, and
  Yang]{nguyen2019hologan}
Thu Nguyen-Phuoc, Chuan Li, Lucas Theis, Christian Richardt, and Yong-Liang
  Yang.
\newblock Holo{GAN}: Unsupervised learning of 3{D} representations from natural
  images.
\newblock In \emph{Proceedings of the IEEE/CVF International Conference on
  Computer Vision}, 2019.

\bibitem[Nguyen-Phuoc et~al.(2020)Nguyen-Phuoc, Richardt, Mai, Yang, and
  Mitra]{blockgan}
Thu Nguyen-Phuoc, Christian Richardt, Long Mai, Yong-Liang Yang, and Niloy
  Mitra.
\newblock Block{GAN}: Learning 3{D} object-aware scene representations from
  unlabelled images.
\newblock \emph{arXiv preprint arXiv:2002.08988}, 2020.

\bibitem[Peters et~al.(2017)Peters, Janzing, and
  Sch{\"o}lkopf]{peters2017elements}
Jonas Peters, Dominik Janzing, and Bernhard Sch{\"o}lkopf.
\newblock \emph{Elements of Causal Inference: Foundations and Learning
  Algorithms}.
\newblock MIT Press, 2017.

\bibitem[Qi et~al.(2017)Qi, Su, Mo, and Guibas]{qi2017pointnet}
Charles~R. Qi, Hao Su, Kaichun Mo, and Leonidas~J. Guibas.
\newblock Point{N}et: Deep learning on point sets for 3{D} classification and
  segmentation.
\newblock In \emph{Proceedings of the IEEE Conference on Computer Vision and
  Pattern Recognition}, 2017.

\bibitem[Rand(1971)]{rand1971}
William~M. Rand.
\newblock Objective criteria for the evaluation of clustering methods.
\newblock \emph{Journal of the American Statistical Association}, 66\penalty0
  (336):\penalty0 846--850, 1971.

\bibitem[Rezende et~al.(2014)Rezende, Mohamed, and Wierstra]{vae_rezende}
Danilo~Jimenez Rezende, Shakir Mohamed, and Daan Wierstra.
\newblock Stochastic backpropagation and approximate inference in deep
  generative models.
\newblock In \emph{International Conference on Machine Learning}, 2014.

\bibitem[Rolls et~al.(2005)Rolls, Xiang, and Franco]{rolls2005object}
Edmund~T. Rolls, Jianzhong Xiang, and Leonardo Franco.
\newblock Object, space, and object-space representations in the primate
  hippocampus.
\newblock \emph{Journal of Neurophysiology}, 94\penalty0 (1):\penalty0
  833--844, 2005.

\bibitem[Sch{\"o}lkopf(2019)]{schlkopf2019causality}
Bernhard Sch{\"o}lkopf.
\newblock Causality for machine learning.
\newblock \emph{arXiv preprint arXiv:1911.10500}, 2019.

\bibitem[Shin et~al.(2019)Shin, Ren, Sudderth, and Fowlkes]{shin20193d}
Daeyun Shin, Zhile Ren, Erik~B. Sudderth, and Charless~C. Fowlkes.
\newblock 3{D} scene reconstruction with multi-layer depth and epipolar
  transformers.
\newblock In \emph{Proceedings of the IEEE/CVF International Conference on
  Computer Vision}, 2019.

\bibitem[Singh et~al.(2019)Singh, Yoon, Son, and Ahn]{snp}
Gautam Singh, Jaesik Yoon, Youngsung Son, and Sungjin Ahn.
\newblock Sequential neural processes.
\newblock In \emph{Advances in Neural Information Processing Systems}, pages
  10254--10264, 2019.

\bibitem[Sitzmann et~al.(2019{\natexlab{a}})Sitzmann, Thies, Heide,
  Nie{\ss}ner, Wetzstein, and Zollhofer]{sitzmann2019deepvoxels}
Vincent Sitzmann, Justus Thies, Felix Heide, Matthias Nie{\ss}ner, Gordon
  Wetzstein, and Michael Zollhofer.
\newblock Deep{V}oxels: Learning persistent 3{D} feature embeddings.
\newblock In \emph{Proceedings of the IEEE/CVF Conference on Computer Vision
  and Pattern Recognition}, 2019{\natexlab{a}}.

\bibitem[Sitzmann et~al.(2019{\natexlab{b}})Sitzmann, Zollh{\"o}fer, and
  Wetzstein]{srn}
Vincent Sitzmann, Michael Zollh{\"o}fer, and Gordon Wetzstein.
\newblock Scene representation networks: Continuous 3{D}-structure-aware neural
  scene representations.
\newblock In \emph{Advances in Neural Information Processing Systems}, pages
  1119--1130, 2019{\natexlab{b}}.

\bibitem[Tobin et~al.(2019)Tobin, Zaremba, and Abbeel]{tobin2019geometry}
Joshua Tobin, Wojciech Zaremba, and Pieter Abbeel.
\newblock Geometry-aware neural rendering.
\newblock In \emph{Advances in Neural Information Processing Systems}, pages
  11555--11565, 2019.

\bibitem[Todorov et~al.(2012)Todorov, Erez, and Tassa]{mujoco}
Emanuel Todorov, Tom Erez, and Yuval Tassa.
\newblock {MuJoCo}: {A} physics engine for model-based control.
\newblock In \emph{2012 IEEE/RSJ International Conference on Intelligent Robots
  and Systems}, pages 5026--5033. IEEE, 2012.

\bibitem[Tulsiani et~al.(2017)Tulsiani, Zhou, Efros, and
  Malik]{tulsiani2017multi}
Shubham Tulsiani, Tinghui Zhou, Alexei~A. Efros, and Jitendra Malik.
\newblock Multi-view supervision for single-view reconstruction via
  differentiable ray consistency.
\newblock In \emph{Proceedings of the IEEE Conference on Computer Vision and
  Pattern Recognition}, 2017.

\bibitem[Tulsiani et~al.(2018)Tulsiani, Gupta, Fouhey, Efros, and
  Malik]{tulsiani2018factoring}
Shubham Tulsiani, Saurabh Gupta, David Fouhey, Alexei~A. Efros, and Jitendra
  Malik.
\newblock Factoring shape, pose, and layout from the 2{D} image of a 3{D}
  scene.
\newblock In \emph{Proceedings of the IEEE/CVF Conference on Computer Vision
  and Pattern Recognition}, 2018.

\bibitem[Tung et~al.(2019)Tung, Cheng, and Fragkiadaki]{grnn}
Hsiao-Yu~Fish Tung, Ricson Cheng, and Katerina Fragkiadaki.
\newblock Learning spatial common sense with geometry-aware recurrent networks.
\newblock In \emph{Proceedings of the IEEE/CVF Conference on Computer Vision
  and Pattern Recognition}, 2019.

\bibitem[van Steenkiste et~al.(2018)van Steenkiste, Kurach, and
  Gelly]{van2018case}
Sjoerd van Steenkiste, Karol Kurach, and Sylvain Gelly.
\newblock A case for object compositionality in deep generative models of
  images.
\newblock \emph{arXiv preprint arXiv:1810.10340}, 2018.

\bibitem[van Steenkiste et~al.(2019)van Steenkiste, Greff, and
  Schmidhuber]{van2019perspective}
Sjoerd van Steenkiste, Klaus Greff, and J{\"u}rgen Schmidhuber.
\newblock A perspective on objects and systematic generalization in model-based
  {RL}.
\newblock \emph{Workshop on Generative Modeling and Model-Based Reasoning for
  Robotics and AI at ICML}, 2019.

\bibitem[van Steenkiste et~al.(2020)van Steenkiste, Kurach, Schmidhuber, and
  Gelly]{van2020investigating}
Sjoerd van Steenkiste, Karol Kurach, J{\"u}rgen Schmidhuber, and Sylvain Gelly.
\newblock Investigating object compositionality in generative adversarial
  networks.
\newblock \emph{Neural Networks}, 130:\penalty0 309--325, 2020.

\bibitem[Veličković et~al.(2018)Veličković, Cucurull, Casanova, Romero,
  Liò, and Bengio]{velivckovic2017graph}
Petar Veličković, Guillem Cucurull, Arantxa Casanova, Adriana Romero, Pietro
  Liò, and Yoshua Bengio.
\newblock Graph attention networks.
\newblock In \emph{International Conference on Learning Representations}, 2018.

\bibitem[von Hofsten and Spelke(1985)]{von1985object}
Claes von Hofsten and Elizabeth~S. Spelke.
\newblock Object perception and object-directed reaching in infancy.
\newblock \emph{Journal of Experimental Psychology: General}, 114\penalty0
  (2):\penalty0 198, 1985.

\bibitem[Watters et~al.(2019)Watters, Matthey, Burgess, and
  Lerchner]{watters2019spatial}
Nicholas Watters, Loic Matthey, Christopher~P. Burgess, and Alexander Lerchner.
\newblock {Spatial Broadcast} decoder: {A} simple architecture for learning
  disentangled representations in {VAE}s.
\newblock \emph{arXiv preprint arXiv:1901.07017}, 2019.

\bibitem[Wu et~al.(2016)Wu, Zhang, Xue, Freeman, and Tenenbaum]{wu2016learning}
Jiajun Wu, Chengkai Zhang, Tianfan Xue, Bill Freeman, and Josh Tenenbaum.
\newblock Learning a probabilistic latent space of object shapes via 3{D}
  generative-adversarial modeling.
\newblock In \emph{Advances in Neural Information Processing Systems}, pages
  82--90, 2016.

\bibitem[Wu and He(2018)]{wu2018group}
Yuxin Wu and Kaiming He.
\newblock Group normalization.
\newblock In \emph{Proceedings of the European Conference on Computer Vision},
  2018.

\bibitem[Yan et~al.(2016)Yan, Yang, Yumer, Guo, and Lee]{yan2016perspective}
Xinchen Yan, Jimei Yang, Ersin Yumer, Yijie Guo, and Honglak Lee.
\newblock Perspective transformer nets: Learning single-view 3{D} object
  reconstruction without 3{D} supervision.
\newblock In \emph{Advances in Neural Information Processing Systems}, pages
  1696--1704, 2016.

\bibitem[Yoon et~al.(2020)Yoon, Singh, and Ahn]{asnp}
Jaesik Yoon, Gautam Singh, and Sungjin Ahn.
\newblock Robustifying sequential neural processes.
\newblock In \emph{International Conference on Machine Learning}, 2020.

\end{thebibliography}

\end{document}